%% file: neurips_2026_inv.tex
\PassOptionsToPackage{numbers,compress}{natbib}
\documentclass{article}

\usepackage[main, final]{neurips_2026}

\usepackage[utf8]{inputenc} 
\usepackage[T1]{fontenc}    
\usepackage{hyperref}       
\usepackage{url}            
\usepackage{booktabs}       
\usepackage{amsfonts}       
\usepackage{nicefrac}       
\usepackage{microtype}      
\usepackage{xcolor}         
\usepackage{graphicx}       
\usepackage{amsmath}
\usepackage{algorithm}
\usepackage{algorithmic}
\usepackage{subcaption}
\usepackage{enumitem}
\usepackage{amssymb}
\usepackage{mathtools}
\usepackage{amsthm}
\usepackage{wrapfig}
\title{Inverse Modeling for Laser Pulse Shape Design in Inertial Confinement Fusion}

\author{%
  Ricardo Luna Gutierrez$^1$$^\dag$, Vineet Gundecha$^1$$^\dag$, Rahman Ejaz$^2$$^\dag$, Varchas Gopalaswamy$^2$$^\dag$, \\ 
  \textbf{Riccardo Betti}$^2$$^\dag$, \textbf{Sahand Ghorbanpour}$^1$$^\dag$, \textbf{Aarne Lees}$^2$$^\dag$, \textbf{Soumyendu Sarkar}$^{1}$$^\dag$\thanks{Corresponding author. \dag These authors contributed equally.} \\
  $^1$Hewlett Packard Enterprise, $^2$University of Rochester \\
  \texttt{\{rluna, vineet.gundecha, sahand.ghorbanpour, soumyendu.sarkar\}@hpe.com} \\
  \texttt{\{reja, vgop, betti, alees\}@lle.rochester.edu}
}

\begin{document}

\maketitle

\begin{abstract}
The achievement of practical fusion energy remains one of the most pressing unsolved scientific challenges, with immense implications for carbon-free power. A critical factor in the success of Inertial Confinement Fusion (ICF) is the design of a Laser Pulse Shape (LP) that can optimally drive implosions under stringent physical constraints. Traditional LP design relies on computationally expensive simulations and manual iterative refinement. We introduce the ICF Laser Pulse Shape Design System (LPDS), a generative inverse modeling framework that maps desired outcomes and target pellet configurations directly to optimized LPs. Crucially, we design a multi-objective loss function to ensure the generated LPs satisfy fundamental physical constraints and experimental requirements. Furthermore, we present constraint-conditioning, inpainting, and gradient-based LP editing mechanisms for fine-grained control over specific pulse characteristics during generation. Moreover, we study the transfer of our framework to a limited set of real experimental shots. Our method establishes a data-driven inverse design framework for LP design in ICF, contributing to the advancement of practical and sustainable fusion energy.
\end{abstract}


\section{Introduction}

Inertial Confinement Fusion (ICF) is a promising route toward fusion energy, but it requires precise design of the Laser Pulse Shape (LP), the temporal profile of laser energy that controls target compression. Crafting effective LPs is particularly challenging because implosion performance depends nonlinearly on small changes in pulse timing, amplitude, and target configuration.  
In the direct-drive setting studied here, LP design relies on repeated radiation-hydrodynamic simulation sweeps and expert iteration with tools such as LILAC \citep{Hansen_2018, LLEReview85, PhysRevA.36.3926, PhysRevLett,Gopalaswamy2019-tc,Gopalaswamy2024-yn}. This forward-design process is computationally expensive and limits rapid exploration of promising designs, creating a need for inverse models that map desired outcomes to candidate LPs.
In this work, we introduce the Laser Pulse Shape Design System (\textbf{LPDS}), a machine learning-based inverse-modeling framework to address this challenge. LPDS enables scalable, data-driven LP design aligned with target implosion objectives and experimental constraints. The main contributions of our work are:  

\begin{itemize}
    \item We propose a unified inverse design framework that jointly integrates closed-loop physics optimization, adaptive constraint reconfiguration, and multi-solution design for ICF.
    \item We propose a constraint-adaptive conditioning mechanism for LP design, enabling rapid reconfiguration to new design constraints without retraining the full generative model.
    \item We formulate a unified multi-objective training scheme combining reconstruction, physics consistency, and energy conservation, allowing controllable trade-offs between diversity and physical fidelity.
    \item We demonstrate generative inverse modeling for LP design in ICF using large-scale simulation data, true LILAC-simulator validation for the best-performing models, and a limited experimental transfer study on 300 experimental shots.
\end{itemize}

\section{Background}

\begin{figure}[t]
    \centering
    \includegraphics[width=\linewidth]{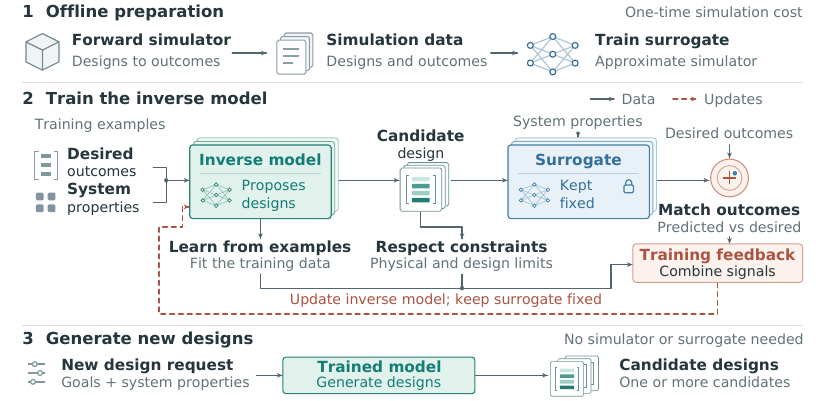}
    \caption{Inverse design workflow. (1) Fit an outcome surrogate to simulations. (2) Train the inverse model using examples, outcome agreement, and constraints; freeze the surrogate. (3) Generate designs from desired outcomes and system properties without simulator or surrogate calls.}
    \label{figure:system_overview}
\end{figure}

ICF achieves nuclear fusion by using high-energy lasers to compress and heat fuel pellets (typically deuterium-tritium) to extreme temperatures and pressures \citep{Betti2016}. Achieving ignition requires precise, symmetric implosion controlled by the LP, the temporal profile governing compression over 3 nanoseconds.

The LP is carefully structured with distinct phases: a low-intensity foot launches timed shocks, a rising ramp controls pressure buildup, a high-intensity main pulse drives rapid implosion, and an optional tail sustains pressure (Figure \ref{fig:sample-pulse}). This multi-phase structure ensures symmetric compression and optimal fusion conditions.

Given the high cost of experiments and limited access to ICF facilities, researchers rely on physics-based simulations to evaluate LP designs. LILAC is a prominent one-dimensional Lagrangian hydrodynamics code for ICF \citep{Hansen_2018, LLEReview85, PhysRevA.36.3926, PhysRevLett,Gopalaswamy2019-tc,Gopalaswamy2024-yn} that maps LPs and target parameters to implosion outcomes such as energy yield, velocity, and areal density. This controlled environment enables systematic exploration of how variations in the LP influence fusion performance for a given target.

LPDS builds on two families of generative models: denoising diffusion models \citep{ho2020denoising, song2020denoising}, and autoregressive sequence models \citep{hochreiter1997long, vaswani2017attention}. Their formal definitions are provided in Appendix~\ref{appendix:background_models}. The iterative denoising process of diffusion models provides a natural mechanism for controlling multi-scale pulse features, while autoregressive models capture the sequential temporal structure of the LP.

\section{Laser Pulse Shape Design System}

We present \textbf{LPDS}, an inverse-modeling framework for generating LPs in ICF. LPDS takes desired implosion outcomes and target pellet parameters as input and generates LPs that satisfy the specified fusion objectives. Figure~\ref{figure:system_overview} summarizes the workflow: offline simulation and surrogate training (1), physics-guided inverse-model training (2), and direct design generation (3). In our ICF application, designs are laser pulses, system properties are pellet parameters, and outcomes are implosion metrics. The following subsections describe each component of LPDS in detail.

\subsection{Inverse Modeling}
Let $\mathbf{m} = \text{LILAC}(\mathbf{l}, \mathbf{p})$ denote the vector of implosion outcomes produced by the LILAC ICF simulator for a given LP ($\mathbf{l}$) and target pellet parameters ($\mathbf{p}$). These outcomes include quantities such as energy yield, areal density, burn width, and ion temperature. The pellet parameters are defined by attributes such as outer radius, ice thickness, ablator thickness, and ice-tritium fraction.

To construct an inverse design model, we first generate a dataset $D_F = \left\{ \left(\mathbf{l}_i, \mathbf{p}_i, \mathbf{m}_i \right) \right\}_{i=1}^I$ of 1 million examples via systematic simulation sweeps, capturing the relationship between pulse shapes, target parameters, and resulting implosion outcomes. The LP $\mathbf{l}_i$ is represented as a real-valued sequence of length 256. $\mathbf{p}_i$ and $\mathbf{m}_i$ are real-valued vectors of size 5 and 12, respectively. The underlying simulation inputs live in a 20-dimensional design space that combines target specifications with an internal parameterization of the pulse shape; to sample high-value regions of this space efficiently, we combine random physically valid draws with perturbations of known high-performing implosions.

Our goal is to learn a data-driven inverse model conditioned on desired implosion outcomes $\mathbf{m}$ and pellet parameters $\mathbf{p}$. For stochastic generators, this corresponds to learning a conditional distribution $p_\theta(\mathbf{l}\mid\mathbf{m}, \mathbf{p})$ from which candidate LPs can be sampled; for deterministic predictors, it corresponds to learning a point estimate $G_\theta(\mathbf{m}, \mathbf{p})$. In the terminology of simulation-based inference, the stochastic variants are amortized neural posterior estimators \citep{cranmer2020frontier}: because LILAC is too expensive to query during training or inference, LPDS learns the posterior from a fixed set of simulations and is then applied to new targets without further simulator calls. For simplicity, we denote a generated candidate pulse by:
\begin{equation}
\mathbf{l}' = G_\theta(\mathbf{m}, \mathbf{p}).
\end{equation}
Conditioning parameters $\mathbf{p}$ and $\mathbf{m}$ are linearly embedded before entering $G_\theta$; Appendix Figure~\ref{fig:lpds_architecture} shows the ICF-specific architecture and optional pulse-shape constraints.

This inverse design problem has multiple plausible solutions, since multiple pulse shapes can lead to similar outcomes. To address this, we consider two modeling approaches:
\begin{itemize}
    \item Generative (Stochastic): Diffusion and autoregressive generative models are trained to sample from a distribution of valid LPs, thereby providing diversity and robustness.
    \item Predictive (Deterministic): We also train predictive autoregressive models to predict a single high-fidelity LP.
\end{itemize}

\subsubsection{Diffusion Model}

Diffusion models iteratively denoise a latent representation to sample from a learned distribution. This property has made them increasingly attractive for physics applications \citep{liu2025diffpic, 10.1609/aaai.v37i4.25663, petersen2023dynamicsdiffusion,shmakov2023endtoend}. 

We condition $G_\theta$ on both the desired implosion outcomes $\mathbf{m}$ and the target parameters $\mathbf{p}$. During the forward diffusion process, noise is gradually added to the original LP $\mathbf{l}$ over $N$ timesteps, resulting in a noisy latent variable $\mathbf{l}_N \sim \mathcal{N}(\mathbf{0}, \mathbf{I})$. The inverse model learns a conditional noise prediction network $\epsilon_\theta(\mathbf{l}_n, n, \mathbf{m}, \mathbf{p})$ that estimates the noise added during a forward process, conditioned on the noisy LP $\mathbf{l}_n$, the timestep $n$, the target implosion outcomes $\mathbf{m}$, and the target parameters $\mathbf{p}$. The training objective of the network $\epsilon_\theta$ is:
\begin{equation}
\mathcal{L}(\theta) = \mathbb{E}_{n\sim [1, N], (\mathbf{l}, \mathbf{m}, \mathbf{p})\sim D_F} \left[ \left\| \epsilon - \epsilon_\theta(\mathbf{l}_n, n, \mathbf{m}, \mathbf{p}) \right\|^2 \right]
\end{equation} 
\begin{equation}
\mathbf{l}_n = \sqrt{\bar{\alpha}_n}\,\mathbf{l} + \sqrt{1-\bar{\alpha}_n}\,\epsilon
\end{equation} 
During the generation phase, we start from a random noise vector $\mathbf{l}_N \sim \mathcal{N}(\mathbf{0}, \mathbf{I})$ and iteratively denoise it using the conditional noise prediction network via the DDIM \citep{song2020denoising} update rule:
\begin{equation}
\mathbf{l}_{n-1} = \sqrt{\bar{\alpha}_{n-1}}\,\mathbf{l}' + \sqrt{1-\bar{\alpha}_{n-1}- \sigma^2_n} \epsilon_\theta(\mathbf{l}_n, n, \mathbf{m}, \mathbf{p}) + \sigma_n\epsilon_n, \label{equation:l_forward}
\end{equation} 
where $\mathbf{l}'$ is estimated via:
\begin{equation}
\mathbf{l}' \approx \frac{\mathbf{l}_n - \sqrt{1-\bar{\alpha}_n}\,\epsilon_\theta(\mathbf{l}_n, n)}{\sqrt{\bar{\alpha}_n}}. \label{equation:x0}
\end{equation}
With this diffusion-based formulation, LPDS generates diverse LPs that remain consistent with target constraints, enabling systematic exploration of physically meaningful designs.

\subsubsection{Autoregressive Model}
\label{AR models}
Alternatively, LP design can be formulated autoregressively, with the pulse vector $\mathbf{l} = \{l_1, l_2, \ldots, l_T\}$ generated sequentially. At each timestep $t$, the model predicts $l_t$ conditioned on the target parameters $\mathbf{p}$, the desired implosion outcomes $\mathbf{m}$, and the previously generated pulse values $\{l_1, \ldots, l_{t-1}\}$. This formulation allows the model to capture the temporal dependencies inherent in the pulse sequence. At the first timestep, when no previous $l_t$ values are available, the model conditions on an initialization vector obtained by projecting the target parameters $\mathbf{p}$ and desired outcomes $\mathbf{m}$ into a shared embedding space. This embedding serves as the initial conditioning input for the autoregressive model.

We explore two paradigms for the autoregressive models: predictive and generative. The predictive models reconstruct the ground-truth LP with high fidelity, whereas the generative models trade off reconstruction fidelity against the ability to provide diverse LP candidates. The training objective for the predictive autoregressive models is the mean squared error between the predicted and true pulse values across all time steps:
\begin{equation}
\mathcal{L}(\theta) = \frac{1}{T} \sum_{t=1}^T \left\| l_t - G_\theta(l_{<t}, \mathbf{m}, \mathbf{p}) \right\|^2,
\end{equation}
where $G_\theta(l_{<t}, \mathbf{m}, \mathbf{p})$ denotes the model's prediction for timestep $t$ given the prior pulse values and conditioning inputs.

At each time step, the generative models output a probability distribution over the LP value at time $t$ denoted by $p_\theta(l_t)$. The training objective for the generative autoregressive models is to minimize the negative log-likelihood of the true pulse under the predicted distribution $p_\theta(l_t)$:

\begin{equation}
\mathcal{L}(\theta) = -\frac{1}{T} \sum_{t=1}^T \log p_{\theta}(l_t \mid l_{<t}, \mathbf{m}, \mathbf{p})
\end{equation}

We explore three modeling choices for $p_\theta$. The simplest models it as a Gaussian distribution, in which $G_\theta$ predicts the mean and variance. To provide greater expressiveness, we also model $p_\theta$ as a weighted mixture of $K$ Gaussians, where the model predicts $K$-dimensional vectors for the means, variances, and mixing weights. 
Finally, we consider a discrete approach in which the pulse $l$ is quantized into 128 values and $p_\theta$ is modeled as a categorical distribution trained with a cross-entropy loss.

\subsection{Auxiliary Objectives}
\label{section:auxiliary_objective}
While the reconstruction objective enables LP reproduction, our broader goal is scientific exploration through plausible, diverse LP generation consistent with implosion outcomes. This provides a wider array of selectable, high-performing candidates and facilitates scientific discovery by potentially identifying novel LP configurations. To achieve this, we integrate a differentiable physics surrogate directly into the generative training loop.

From an optimization perspective, this process can be viewed as a bilevel problem in which a generative model proposes candidate designs while a physics model evaluates their feasibility and performance. LPDS approximates this process by optimizing the inverse model end to end through a frozen differentiable surrogate, directly aligning the generative distribution with the physically valid solution manifold. By incorporating differentiable physics surrogates into the training loop, the model does not merely condition on observed data but actively optimizes generated designs toward target physical outcomes.

For this purpose, we train a surrogate model\footnote{The surrogate achieves an average error of 1.4\% across the implosion outcomes $\mathbf{m}$ (details in Appendix \ref{section:LILAC surrogate}). This should be interpreted as the error floor of the surrogate-based evaluation, which is why we also report true LILAC-simulator validation for the strongest models.} $\mathbf{S}_\phi$ to approximate the LILAC simulator. This model learns the mapping from an LP $\mathbf{l}$ and target parameters $\mathbf{p}$ to the corresponding implosion outcomes $\mathbf{m}$. This can be represented as $\hat{\mathbf{m}} = \mathbf{S}_\phi(\mathbf{l}, \mathbf{p})$, where $\phi$ represents the learnable parameters of the surrogate model. The surrogate is trained on the simulation dataset $D_F$ by minimizing the prediction error:
\begin{equation}
\text{LILAC}_{\text{surrogate}}(\phi) = \mathbb{E}_{(\mathbf{l}, \mathbf{m}, \mathbf{p})\sim D_F} \left[ \left\| \mathbf{m} - \mathbf{S}_\phi(\mathbf{l}, \mathbf{p}) \right\|^2 \right].
\end{equation}
After training, $\mathbf{S}_\phi $ is kept frozen and is used during training of the inverse model to evaluate the physical fidelity of generated LPs. Given a generated pulse $ \mathbf{l}' = G_\theta(\mathbf{m}, \mathbf{p}) $, the surrogate model predicts the corresponding outcomes $ \hat{\mathbf{m}} = \mathbf{S}_\phi(\mathbf{l}', \mathbf{p}) $, which are then compared to the target outcomes $\mathbf{m}$. Our new objective can be defined as:
\begin{align}
\mathcal{O}(\theta) &= \lambda_{\mathcal{L}} \mathcal{L}(\theta) + \lambda_{\mathcal{S}} \mathcal{S}(\theta) \label{eq:generator_objective} \\
\mathcal{S}(\theta) &= \mathbb{E}_{\mathbf{l}' \sim G_\theta, (\mathbf{m}, \mathbf{p})\sim D_F} \left[ \left\| \mathbf{m} - \mathbf{S}_\phi(\mathbf{l'}, \mathbf{p}) \right\|^2 \right] \label{eq:surrogate_loss_in_generator}
\end{align}
Here, $\lambda_{\mathcal{L}}$ and $\lambda_{\mathcal{S}}$ control the trade-off between faithfully reconstructing the original LP and generating physically consistent alternatives. Properly tuning these weights allows the model to generalize beyond pure reconstruction, producing diverse yet plausible LPs tailored to the specified design goals. In practice, we first train only on the reconstruction loss ${\mathcal{L}(\theta)}$ since at the beginning of training $G_\theta$ can produce LPs that are incoherent and out-of-distribution for $S_\phi$.

For our diffusion model, this loss formulation $\mathcal{S}(\theta)$ appears to require knowledge of the clean LP at each reverse step. However, DDIM's non-Markovian formulation allows us to estimate $\hat{\mathbf{l}}_0$ at any intermediate timestep $n$ using Equation \ref{equation:x0}.

\subsection{Physics-Informed Loss}

While our goal is to generate a diverse set of LPs for a given target configuration, the practical utility of these designs depends critically on their adherence to the fundamental physics of ICF and their feasibility under real-world experimental constraints. One essential constraint is energy conservation. Specifically, generated LPs should not exceed the total energy budget of corresponding reference pulses. For any \( \mathbf{l}_i \in D_F \), we impose the following constraint on a generated LP \( \mathbf{l}'_i \):
\[
\int_0^T \mathbf{l}'_i(t) \, dt \leq \int_0^T \mathbf{l}_i(t) \, dt, \quad t \in [0, T],
\]
where \( T \) is the total pulse duration. To enforce this, we introduce a physics-based penalty term:
\begin{equation}
\mathcal{P}(\theta) = \mathbb{E}_{\mathbf{l}' \sim G_\theta, \mathbf{l} \sim D_F}  
\left[ \left( \int_0^T \mathbf{l}'(t) \, dt - \int_0^T \mathbf{l}(t) \, dt \right)^+ \right]
\end{equation}
This term penalizes excess energy, encouraging designs within experimental and physical limits. Incorporating this constraint into our overall loss function, we define the complete training objective as:
\begin{equation}
\mathcal{O}(\theta) = \lambda_{\mathcal{L}} \mathcal{L}(\theta) + \lambda_{\mathcal{S}} \mathcal{S}(\theta) + \lambda_{\mathcal{P}} \mathcal{P}(\theta) \label{eq:composite_equation}
\end{equation}
where $\lambda_{\mathcal{L}}$, $\lambda_{\mathcal{S}}$, and $\lambda_{\mathcal{P}}$ are hyperparameters that control the relative importance of each loss term in the overall objective (Ablation of $\lambda$ in Appendix Section \ref{sec:fid-div}). This joint objective enables the generation of LPs that are diverse, outcome-consistent, physically plausible, and experimentally viable.

\section{Experiments}

\subsection{Performance Evaluation}
\label{section:performance_evaluation}

We compare \textbf{LPDS} variants across deterministic models (an autoregressive Transformer \citep{vaswani2017attention}, $\mathbf{LPDS}_\text{Transformer}$, and an LSTM \citep{hochreiter1997long}, $\mathbf{LPDS}_\text{LSTM}$) and generative models (a diffusion model, $\mathbf{LPDS}_\text{Diff}$, and three autoregressive LSTM variants with Gaussian, Mixture-of-Gaussians, and categorical output distributions). A comparison against seven off-the-shelf conditional generators and inverse-problem methods is reported in Appendix~\ref{appendix:offtheshelf_baselines}, and generalization to unseen pellet configurations is evaluated in Appendix~\ref{sec:pellet_stratified}.

We also perform an ablation study to assess the impact of our proposed objectives by evaluating versions of the model without the surrogate loss $\mathcal{S}(\theta)$ ($w/o\ \mathcal{S}$) and without the physics-informed loss $\mathcal{P}(\theta)$ ($w/o\ \mathcal{P}$).

We report four key metrics: implosion-outcome error in $\mathbf{m}$, reconstruction error, generation diversity, and energy-conservation violation. All models were evaluated over $5$ full training runs and $R=10$ random seeds (50 runs per model) on a test set $D_{F_{\text{test}}} = \left\{ \left(\mathbf{l}_e, \mathbf{p}_e, \mathbf{m}_e \right) \right\}_{e=1}^E$, where $D_{F_{\text{test}}} \cap D_F = \emptyset$.

The implosion-outcome error in $\mathbf{m}$ is computed via the surrogate loss (Equation~\ref{eq:surrogate_loss_in_generator}) and reported as mean absolute percentage error (MAPE). Reconstruction error is computed as $\frac{1}{E} \sum_{e=1}^{E} \frac{1}{R} \sum_{r=1}^{R} \left\| \mathbf{l}_e - \mathbf{l}'_{e,r} \right\|$. Energy-conservation violation is reported as the mean positive excess-energy percentage, $\frac{1}{E} \sum_{e=1}^{E} \frac{1}{R} \sum_{r=1}^{R} \frac{\left( \int_0^J \mathbf{l}'_{e,r}(j) \, dj - \int_0^J \mathbf{l}_e(j) \, dj \right)^+}{\int_0^J \mathbf{l}_e(j) \, dj} \times 100$, so lower values indicate fewer budget violations. Diversity is measured by average pairwise $L_2$ distance across generated samples: $\frac{1}{E} \sum_{e=1}^{E} \frac{1}{\binom{R}{2}} \sum_{1 \leq j < k \leq R} \left\| \mathbf{l}'_{j,e} - \mathbf{l}'_{k,e} \right\|_2$. The estimated upper bound for diversity is $1.9$, obtained when comparing randomly selected LPs from $D_{F_{\text{test}}}$.

We also evaluate the implosion-outcome error on the \textbf{true LILAC simulator}, rather than relying only on the surrogate, and report this quantity as the \textbf{LILAC error}. Because simulator access is limited, this validation is performed for the two best-performing models ($\mathbf{LPDS}_\text{LSTM}$ and $\mathbf{LPDS}_\text{Diff}$). The LILAC error therefore serves as the primary simulator-grounded validation of LPDS and should not be conflated with surrogate-based screening.

Table~\ref{table:experimental_results} summarizes the main results. The ablation studies confirm the critical roles of $\mathcal{S}(\theta)$ and $\mathcal{P}(\theta)$. Among LPDS variants, $\mathbf{LPDS}_\text{LSTM}$ achieves the highest reconstruction fidelity but yields zero diversity. The generative variants ($\mathbf{LPDS}_\text{Diff}$ and $\mathbf{LPDS}_\text{*AR}$) maintain low implosion-outcome error while producing varied, high-quality pulses (Figure~\ref{fig:overview}a), with $\mathbf{LPDS}_\text{Diff}$ providing the greatest diversity while preserving low $\mathbf{m}$ error. This precision--diversity trade-off enables scientists to select predictive models for exact replication or generative models for constraint-driven exploration; example generations are shown in Appendix Figures~\ref{figure:ps_ours}--\ref{figure:ps-mixture_gaussians}. The two best variants ($\mathbf{LPDS}_\text{Diff}$ and $\mathbf{LPDS}_\text{LSTM}$) both achieve less than $2.1\%$ error on the true LILAC simulator, confirming that surrogate-based gains transfer to the simulator. Per-metric bar plots of these results are provided in Appendix~\ref{appendix:offtheshelf_baselines} (Figures~\ref{fig:bar_main} and~\ref{fig:bar_ablation}).

\begin{table*}[h]
\centering
\resizebox{\textwidth}{!}{%
\begin{tabular}{lccccc}
\toprule
\textbf{LPDS variant} & \textbf{Diversity} $\uparrow$ & $\mathbf{m}$\textbf{ Error} $\downarrow$ & \textbf{Reconstruction} $\downarrow$ & \textbf{Energy Cons.} $\downarrow$ & \textbf{LILAC Error} $\downarrow$\\
 & & (\%) & ($\times 10^{-3}$) & (\%) & (\%) \\
\midrule
Diff        & $0.64\pm0.077$ & $1.93\pm0.046$ & $5.6\pm0.20$ & $1.59\pm0.14$  & $2.07$ \\
LSTM        & --              & $1.65\pm0.09$  & $0.1\pm0.05$ & $0.66\pm0.07$  & $2.01$ \\
Transformer & --              & $1.94\pm0.11$  & $0.8\pm0.08$ & $0.95\pm0.10$  & -- \\
LSTM$_{\text{Gaussian}}$          & $0.42\pm0.02$  & $1.89\pm0.01$  & $0.5\pm0.02$ & $0.58\pm0.004$ & -- \\
LSTM$_{\text{MixtureOfGaussian}}$ & $0.56\pm0.10$  & $1.95\pm0.09$  & $0.6\pm0.08$ & $1.58\pm0.006$ & -- \\
LSTM$_{\text{Categorical}}$       & $0.39\pm0.04$  & $2.01\pm0.04$  & $0.9\pm0.05$ & $1.18\pm0.08$  & -- \\
\midrule
Diff w/o $\mathcal{S}$        & $0.62\pm0.069$ & $4.50\pm0.014$ & $5.7\pm0.09$ & $1.71\pm0.005$ & -- \\
LSTM w/o $\mathcal{S}$        & --              & $3.90\pm0.011$ & $0.4\pm0.01$ & $0.85\pm0.001$ & -- \\
Transformer w/o $\mathcal{S}$ & --              & $4.40\pm0.016$ & $1.0\pm0.02$ & $0.95\pm0.002$ & -- \\
\midrule
Diff w/o $\mathcal{P}$        & $0.69\pm0.084$ & $1.99\pm0.010$ & $7.4\pm0.10$ & $3.21\pm0.007$ & -- \\
LSTM w/o $\mathcal{P}$        & --              & $1.85\pm0.010$ & $0.1\pm0.05$ & $0.78\pm0.001$ & -- \\
Transformer w/o $\mathcal{P}$ & --              & $2.10\pm0.010$ & $0.9\pm0.07$ & $0.94\pm0.002$ & -- \\
\bottomrule
\end{tabular}
}
\caption{ICF model performance. $\pm$ denotes standard deviation over seeds and training runs. For the predictive autoregressive models ($\mathbf{LPDS}_\text{LSTM}$ and $\mathbf{LPDS}_\text{Transformer}$), diversity is not defined because they are deterministic. The LILAC-error column reports true simulator validation rather than surrogate evaluation and is available only for the strongest models because simulator access is limited. The ablation experiments for the loss terms were performed for a subset of the models due to computational constraints.}
\label{table:experimental_results}
\end{table*}

\section{Pulse Shape Constrained Design}
\label{pulse-shape-constraining}
\begin{wrapfigure}{r}{0.5\textwidth}
    \centering
    \includegraphics[width=\linewidth]{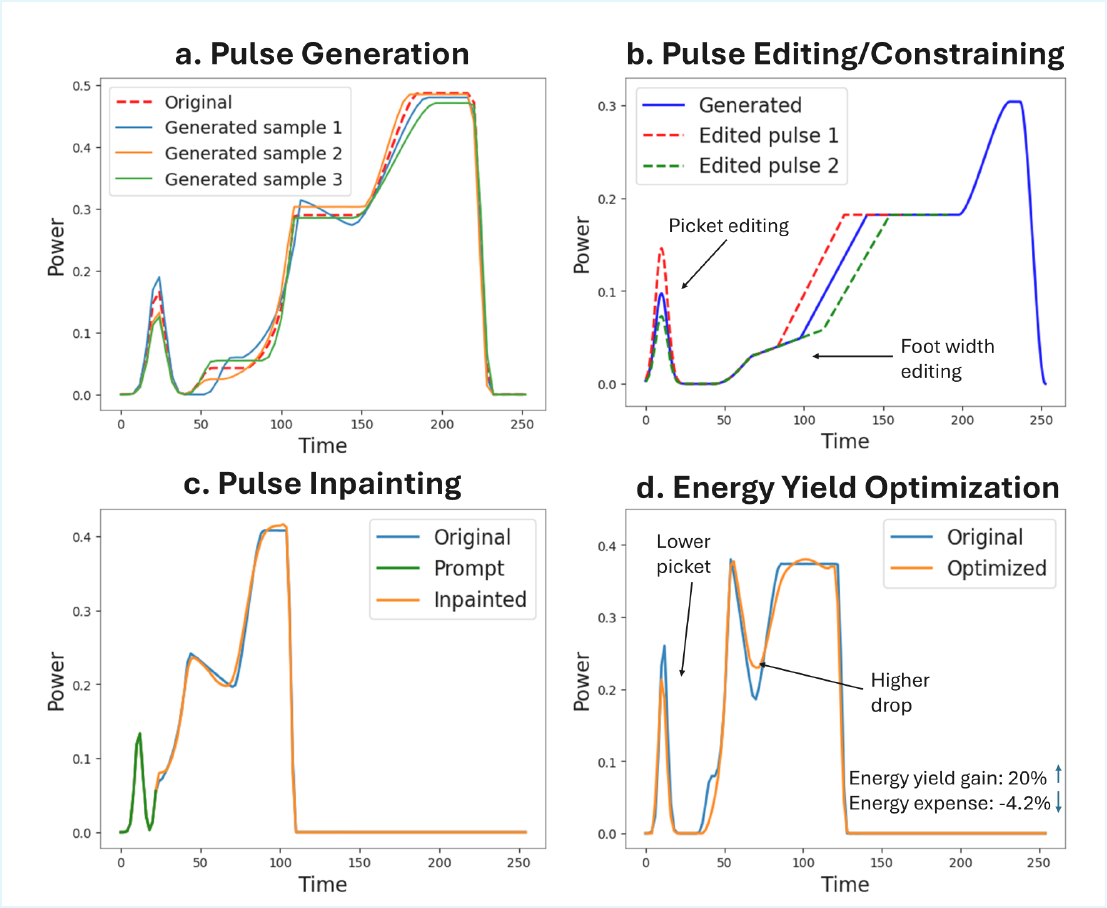}
    \caption{Overview of LPDS capabilities. Panel (a) shows diverse conditional generation of valid pulse shapes, panel (b) shows constraint-based editing of specific pulse attributes, panel (c) shows inpainting of partially specified pulses, and panel (d) shows outcome-driven optimization followed by simulator validation.}
    \label{fig:overview}
\end{wrapfigure}

Figure~\ref{fig:overview} illustrates the generation, editing, inpainting, and optimization capabilities supported by LPDS.

In many ICF design scenarios, scientists require control over specific regions or attributes of the LP. We use a mapping function $\mathcal{M}$, implemented as an LSTM trained on the same dataset (details in Appendix \ref{section:mapping_function}), to project an LP $\mathbf{l}$ onto $M = 12$ parameters $C = \mathcal{M}(\mathbf{l}) = \{c^1, \dots, c^{12}\}$. Each $c^m$ corresponds to a physically interpretable property of the LP (Figure \ref{fig:sample-pulse}). When constructing new designs, scientists may fix one or two of these values while allowing the remaining parameters, and thus the overall pulse shape, to vary.

However, our current inverse model $G_\theta$ lacks a mechanism to enforce such partial constraints during inference. A straightforward solution would be to take the constraint parameter $\mathbf{c}^{m}$ as an additional input during training, producing an LP conditioned on the desired implosion outcomes $\mathbf{m}$, target parameters $\mathbf{p}$, and the specified constraint as $\mathbf{l}' = G_\theta(\mathbf{m}, \mathbf{p}, \mathbf{c}^m)$.
A constraining loss can then be introduced to enforce consistency between the specified constraint values and their corresponding values in the generated LP:
\begin{equation}
\label{equation:j_theta}
\mathcal{J}(\theta) =\mathbb{E}_{\mathbf{l}', c^m} \left[ \left\| c^m_{l'} - c^m \right\|^2 \right]
\end{equation}
where $c^m_{l'} \in C_{l'}, C_{l'} = \mathcal{M}(\mathbf{l}')$. The constrained $G_\theta$ objective can be defined as:
\begin{equation}
\label{equation:constraining}
\mathcal{O}(\theta) = \lambda_{\mathcal{L}} \mathcal{L}(\theta) + \lambda_{\mathcal{S}} \mathcal{S}(\theta) + \lambda_{\mathcal{P}} \mathcal{P}(\theta) + \lambda_{\mathcal{J}} \mathcal{J}(\theta)
\end{equation}

\subsection{Diffusion-Based Adaptation}
\label{sec:diffusion_adaptation}

The above approach enables $G_\theta$ to satisfy the specified constraints. However, exhaustively training separate diffusion models for all combinations of one or two parameters in $C$ is computationally infeasible due to combinatorial complexity.

To address this challenge, we propose a rapid adaptation strategy ~\citep{pmlr-v84-bartunov18a, netanyahu2025fewshottasklearninginverse, giannone2022fewshotdiffusionmodels, hu2024anomalydiffusion}. Rather than retraining from scratch, we adapt a pre-trained model $G_\theta(\mathbf{m},\mathbf{p})$ with a small number of gradient updates to satisfy a new constraint $c$.

Our adaptation approach employs an embedding network $\mathcal{B}_\delta$ that encodes constraint parameters $c \in C$ to condition LP generation. This results in a conditioned noise model: $\epsilon_\theta(\mathbf{l}_n, n, \mathbf{m}, \mathbf{p}, \mathcal{B}_\delta(c))$. To simplify conditioning, we enforce equal embedding dimensionality for $n$ and $c = \mathcal{B}_\delta(c)$, enabling their combination into a single conditioning variable $u = n + c$. This allows us to condition the model as $\epsilon_\theta(\mathbf{l}_t, u, \mathbf{m}, \mathbf{p})$, requiring no architectural changes to the original model. During adaptation, only the parameters of $\mathcal{B}_\delta$ are fine-tuned, according to Equation \ref{equation:constraining}, while the remaining weights are frozen. 

Conceptually, the proposed adaptation mechanism can be interpreted as learning a low-dimensional constraint embedding that shifts the generative distribution toward constraint-satisfying submanifolds. Rather than retraining the full generative model for each new constraint configuration, LPDS performs lightweight updates that reconfigure the generative process while preserving the learned global structure of the design space.

Examples of this adaptation technique for picket power are shown in Figure \ref{figure:pulse_constrain_diffusion} (Appendix). For a given constraint $c$, the model is fine-tuned once and can then generate an arbitrary number of pulses for different specifications of $\mathbf{m}$ and $\mathbf{p}$.

\subsection{Gradient-Based Adaptation}
\label{sec:grad_adaptation}
Alternatively, the LP can be adapted after generation using a model-agnostic technique without additional fine-tuning. To honor constraints in $\mathcal{C}$ while maintaining desired implosion outcomes, a loss combining the pulse constraining loss $\mathcal{J(\theta)}$ (Equation \ref{equation:j_theta}) and the LILAC surrogate loss $\mathcal{S(\theta)}$ (Equation \ref{eq:surrogate_loss_in_generator}) is formulated. Since both losses are differentiable w.r.t. the pulse $\mathbf{l}'$ (now referred to as $\mathcal{J(\mathbf{l}')}$ and $\mathcal{S(\mathbf{l}')}$), we can perform gradient descent with respect to the following loss and update the pulse $\mathbf{l}'$:
\begin{equation}
\mathcal{T}(\mathbf{l}') = \mathcal{S}(\mathbf{l}') + \lambda_{\mathcal{J}} \mathcal{J}(\mathbf{l}')
\end{equation}
\begin{equation}
\mathbf{l}' := \mathbf{l}' - \alpha \nabla_{\mathbf{l}'} \mathcal{T}(\mathbf{l}')
\end{equation}
where $\alpha$ is the learning rate. One drawback of this technique is that gradient descent can push the LP into undesirable regions, such as negative values or peaks that violate the maximum power. In practice, we mitigate these issues by clipping the pulse. In addition, gradient-based adaptation must be performed separately for each constrained LP, whereas our diffusion-based adaptation produces a fine-tuned model that can be reused indefinitely.

\subsection{LP Constraining Performance}
We evaluate the ability of \textbf{LPDS} to rapidly adapt to specific design constraints $c$, using both diffusion-based and gradient-based adaptation strategies. For the diffusion-based approach, the model is fine-tuned for only 10 epochs on a new constraint parameter $c^m$, requiring minimal computational overhead. In parallel, we apply the gradient-based adaptation to the deterministic model $\text{LPDS}_\text{LSTM}$, allowing for direct post-generation refinement.

We focus our evaluation on two physically meaningful parameters: picket power and foot power (see Figure~\ref{fig:sample-pulse}). To quantify adaptation accuracy, we compute the mean absolute percentage error (MAPE) between the specified constraint value $c$ and the corresponding value extracted from the generated pulse (as defined in Equation~\ref{equation:j_theta}). The gradient-based approach achieves a MAPE of $\mathbf{2.5\%\pm0.12}$, while the diffusion-based method yields a MAPE of $\mathbf{3.4\%\pm0.14}$. Importantly, both adaptation methods preserve LPDS’s performance on the target implosion outcomes ($\mathbf{m}$\textbf{-Error} remains unaffected).

These constraint adaptation strategies empower scientists to generate diverse, high-quality pulse shapes that meet desired implosion outcomes and can be interactively tuned to satisfy evolving experimental or engineering constraints.


    

\subsection{Inpainting/Prompting}
Inpainting provides control directly in LP space: users specify a prefix or fixed regions rather than parameters in $\mathcal{C}$. Autoregressive models complete the pulse from a supplied prefix, while diffusion conditions denoising on specified segments. In our evaluation, the autoregressive prompt covers the first 10\% of the pulse (the first peak); diffusion receives random observed segments with the remainder masked. Diffusion and LSTM achieve $\mathbf{m}$ errors of $(2.00\pm0.012)\%$ and $(1.55\pm0.010)\%$, respectively, with reconstruction errors of $(2.0\pm0.2)\times10^{-5}$ and $(0.3\pm0.02)\times10^{-5}$ ($\pm$ denotes standard deviation over seeds). Appendix Figures~\ref{figure:inpainting_lps} and~\ref{figure:inpainting_ar_lps} show example completions.

\subsection{Novel LP Design for Energy Yield Optimization}
\label{section:optimization}

Beyond reconstructing existing pulse shapes, \textbf{LPDS} can discover novel laser pulse designs that improve implosion performance under strict physical constraints. We formulate pulse optimization as a constrained generative inverse-design problem in which the goal is to generate new pulse shapes that outperform high-quality baseline designs while preserving energy feasibility.

Specifically, given an initial high-performing pulse $\mathbf{l}_e$ with system parameters $\mathbf{p}_e$ and measured implosion outcomes $\mathbf{m}_e$, we define a target outcome by increasing the energy yield component by 10\%:
\[
\tilde{\mathbf{m}}_{e}^{\text{yield}} = 1.10 \times \mathbf{m}_{e}^{\text{yield}}.
\]
Using this target, LPDS generates candidate optimized pulses via
\[
\tilde{\mathbf{l}}'_e = G_\theta(\mathbf{p}_e, \tilde{\mathbf{m}}_e).
\]

The generated pulse must satisfy two constraints: (i) achieving a predicted energy yield at least as high as the target and (ii) maintaining the original total energy budget,
\vspace{-0.1in}
\[
\mathbf{S}_\phi(\tilde{\mathbf{l}}'_e)^{\text{yield}} \geq \tilde{\mathbf{m}}_{e}^{\text{yield}}, \quad 
\int_0^J \tilde{\mathbf{l}}'_e(j)\, dj \leq \int_0^J \mathbf{l}_e(j)\, dj.
\]

Here, $\mathbf{S}_\phi$ denotes the surrogate model used to estimate implosion performance. Since the generated LP may lie beyond the physically viable distribution, we further impose a frequency-domain similarity constraint to ensure the generated LPs are physically plausible:
\[
\frac{1}{J} \sum_{j=1}^{J} \left|\, |\mathcal{F}(\mathbf{l}_e)_j| - |\mathcal{F}(\tilde{\mathbf{l}}'_e)_j| \,\right| < \tau,
\]
where $\mathcal{F}$ is the discrete Fourier transform, $J$ is the number of temporal samples and $\tau$ controls the allowable structural deviation. This constraint ensures that LPDS explores novel yet physically realistic pulse variations within the learned design manifold.

For each initial design, we generate 20 candidate pulses using different random seeds and select the pulse that maximizes predicted energy gain while satisfying all constraints. Crucially, the final optimization results are not reported through the surrogate: the selected optimized pulses are evaluated with the true \textbf{LILAC hydrodynamic simulator} on 60 unique high-performing baseline pulses.

We compare the diffusion-based generator $\mathbf{LPDS}_{\text{Diff}}$ with the autoregressive variant $\mathbf{LPDS}_{\text{LSTM}}$. Under true LILAC validation, the autoregressive model achieves a modest mean energy-yield improvement of \textbf{2.79\%}. In contrast, $\mathbf{LPDS}_{\text{Diff}}$ consistently discovers novel pulse structures that produce a substantially higher mean gain of \textbf{17.74\%} while respecting the original energy budget, and \textbf{86\%} of the optimized pulses exceed the 10\% yield-improvement target. A mean gain of $17.74\%$ corresponds to $\sim$1.18$\times$ neutron yield, obtained on top of baseline pulses that are already high-performing. These results demonstrate that LPDS can effectively explore the constrained design space and uncover superior pulse configurations beyond the baseline designs.

\section{Limited Experimental Transfer Study}

Real-world ICF data are scarce because experiments are expensive and limited in number. To assess transferability beyond simulation, we conduct a limited transfer study on historical experimental data; this is a retrospective evaluation rather than a prospective facility experiment. We fine-tune \textbf{LPDS} on 300 experimental laser pulses, comprising the complete set recorded to date, using a 70\%--30\% train--test split. Despite the limited sample size, measurement noise, and domain shift, the surrogate $\mathbf{S_\phi}$ achieves 8\% test prediction error. On the experimental test set, \textbf{LPDS} achieves:
\begin{itemize}[itemsep=1pt, parsep=0pt, topsep=0pt]
    \item an \textbf{implosion-outcome error} of $8.8\% \pm 0.05$,
    \item a \textbf{reconstruction error} of $0.0068 \pm 0.0001$, and
    \item an \textbf{energy conservation error} of $1.78\% \pm 0.006$.
\end{itemize}
These results are broadly consistent with simulation performance, particularly in reconstruction quality and physical plausibility. The higher $\mathbf{m}$\textbf{-Error} is expected given experimental noise and surrogate limitations in this regime. Overall, LPDS shows encouraging transfer from simulation to experiment with minimal fine-tuning; prospective validation on new facility shots remains future work.

\section{Related Work}

\paragraph{Inverse modeling and simulation-based inference.}
ML-based inverse modeling spans diffusion models \cite{daras2024surveydiffusionmodelsinverse, chung2023diffusion, NEURIPS2021_7d6044e9, song2023pseudoinverseguided}, autoregressive sequence models \citep{duthe2025graphtransformersinversephysics, Luo_2022, 10.1145/3474085.3475436}, GANs and VAEs \citep{ray2021efficient, anirudh2018unsupervisedapproachsolvinginverse, Kingma2013, Liu2021}, and other paradigms \citep{Wang_2021, kamyab2021surveydeeplearningmethods, Adler_2017}.  In the physical sciences, inverse methods have been applied to thermodynamics \citep{ISHITSUKA2023120855}, materials discovery \citep{TAMADDONJAHROMI2020113217, HAGHIGHAT2021113741}, geophysics \citep{10.1093/gji/ggz204}, and high-energy/plasma physics \citep{shmakov2023endtoend, ferreira2020deep, PhysRevD.109.012010, ozturk2024inverserenderingfusionplasmas}.  However, these methods are typically domain-specific and focus on reconstructive inverse problems under fixed forward operators, producing single or narrowly distributed solutions. Simulation-based inference \citep{cranmer2020frontier} provides a general framework for amortized posterior estimation with expensive simulators, but neural posterior estimators are typically applied to low-dimensional parameter vectors and do not natively address three requirements of LP design: a 256-dimensional structured output, hard physical constraints that must hold for every sample, and interactive control through inpainting and editing. Flow matching \citep{lipman2023flow} is a further generative family that could in principle be combined with the surrogate coupling, energy penalty, and inpainting used in LPDS; we leave its evaluation to future work.

\paragraph{Data-driven modeling for direct-drive ICF.}
Statistical models calibrated on experimental data have substantially improved direct-drive implosion performance, most notably tripling the fusion yield on OMEGA \citep{Gopalaswamy2019-tc} and enabling a hydrodynamically equivalent burning plasma \citep{Gopalaswamy2024-yn}; related work identifies the experimental dependencies of fusion yield on implosion parameters \citep{PhysRevLett}. Machine learning has also been applied to fusion plasmas for tomography, inverse rendering, and accelerated simulation \citep{ferreira2020deep, ozturk2024inverserenderingfusionplasmas, liu2025diffpic}. These works address forward modeling, diagnostics, or simulation acceleration; LPDS addresses the inverse direction, learning a conditional generative model that proposes multiple pulse shapes for desired outcomes and supports controllable editing and simulator-validated optimization. To our knowledge, no prior work develops such a model for the direct-drive LP design setting studied here.

\paragraph{Optimization of expensive simulators.}
When evaluating the objective requires an expensive simulator, Bayesian optimization \citep{snoek2012practical} and evolution strategies such as CMA-ES \citep{hansen2001cmaes} are standard black-box optimizers. These methods solve one design problem at a time and must be rerun for every new target, whereas LPDS amortizes the inverse mapping across targets and pellet configurations. Direct optimization over a low-dimensional pulse parameterization against our surrogate is evaluated in Appendix~\ref{appendix:constrained_optimization}.

\section{Broader Applicability}

Although LPDS is instantiated for laser-pulse design in ICF, the framework is applicable to a broader class of scientific and engineering inverse-design problems. Many domains, including materials discovery, climate-aware infrastructure control, chemical process design, and accelerator tuning, require generating high-dimensional inputs that satisfy desired outcomes while obeying physical, operational, or safety constraints. The core LPDS ingredients, including conditional generative modeling, differentiable surrogate feedback, constraint-aware penalties, and post-generation editing, do not depend on fusion-specific assumptions. They can be reused whenever a simulator, experimental archive, or learned forward model can evaluate candidate designs, enabling diverse and controllable solution generation beyond settings where a single deterministic inverse mapping is sufficient.

\section{Conclusion and Limitations}
\label{section:conclusion}
We introduced LPDS, an inverse-design framework that integrates conditional generative models with physics-guided optimization and adaptive constraint control for LP shaping in ICF. LPDS enables multi-solution inverse modeling while enforcing physical feasibility through differentiable surrogate models and a physics-informed loss. Experiments on simulated data, true LILAC-simulator validation, and a limited experimental transfer study demonstrate accurate reconstruction, flexible constraint handling, and the discovery of novel pulse designs that improve energy yield.

\paragraph{Limitations}

The method depends on a learned surrogate during training and most large-scale evaluation, so surrogate bias remains a central limitation. In addition, the real experimental study is necessarily small, and the experimental data are noisier and less abundant than the simulation data. Finally, although the energy-conservation term enforces an important physical constraint, it does not by itself guarantee full facility-level experimental feasibility or robustness under out-of-distribution target requests. Regarding safety, a mis-designed pulse in direct-drive ICF degrades implosion performance rather than creating a runaway condition; generated pulses are additionally screened by the energy-conservation and frequency-domain constraints and, like any new design, would undergo the standard facility safety review before being fielded.

\section*{Acknowledgments} This work was supported by the U.S. Department of Energy, Office of Science, under Award Nos. DE-SC0024408 and DE-SC0024381.


\bibliographystyle{plainnat}
\bibliography{ref}


\newpage
\appendix
\onecolumn

\section{Background: Generative Models}
\label{appendix:background_models}

\textbf{Denoising Diffusion Probabilistic Models (DDPMs).}
DDPMs \citep{ho2020denoising} define a forward process that gradually adds Gaussian noise to data $\mathbf{x}_0 \sim q(\mathbf{x}_0)$ over $T$ steps, producing increasingly noisy latents $\mathbf{x}_1, \dots, \mathbf{x}_T$.  The transition is $q(\mathbf{x}_t | \mathbf{x}_{t-1}) = \mathcal{N}(\mathbf{x}_t;\, \sqrt{1-\beta_t}\,\mathbf{x}_{t-1},\, \beta_t \mathbf{I})$, where $\{\beta_t\}$ is a fixed variance schedule.  A neural network $\epsilon_\theta(\mathbf{x}_t, t)$ is trained to predict the added noise:
\[
\mathcal{L}_{\text{DDPM}} = \mathbb{E}_{t, \mathbf{x}_0, \boldsymbol{\epsilon}} \bigl[\|\boldsymbol{\epsilon} - \epsilon_\theta(\mathbf{x}_t, t)\|^2 \bigr].
\]
Sampling proceeds by iteratively denoising from $\mathbf{x}_T \sim \mathcal{N}(\mathbf{0}, \mathbf{I})$, recovering clean data through the learned reverse process.

\textbf{Autoregressive (AR) Models.}
AR models \citep{hochreiter1997long, vaswani2017attention} decompose the joint distribution as $p(\mathbf{x}) = \prod_{j=1}^{J} p(x_j \mid x_{<j})$, generating each element conditioned on all preceding ones.  This sequential factorization naturally captures the temporal structure of LPs and enforces causal ordering.

\section{Experimental Details}
\label{section:experimental details}

\begin{figure}[tbp]
  \centering
  \includegraphics[width=0.9\linewidth]{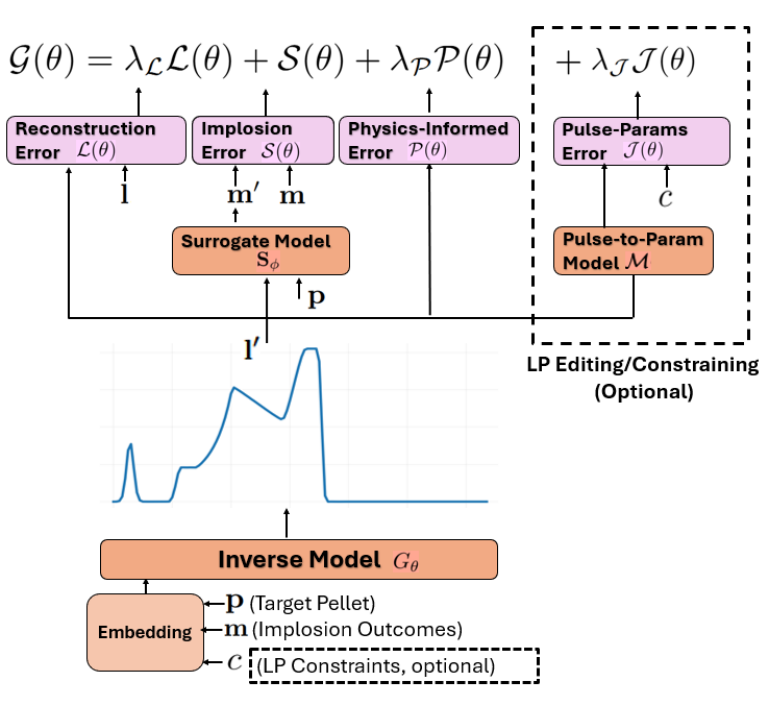}
  \caption{ICF-specific LPDS architecture. The inverse model $G_\theta$ generates a laser pulse $\mathbf{l}'$ from embedded pellet parameters $\mathbf{p}$ and desired implosion outcomes $\mathbf{m}$. Training combines pulse reconstruction, surrogate-predicted outcome agreement, and physics-informed penalties. The dashed branch adds optional pulse-shape constraints $\mathbf{c}$ through a pulse-to-parameter model $\mathcal{M}$.}
  \label{fig:lpds_architecture}
\end{figure}

\begin{figure}[tbp]
  \centering
    
  \includegraphics[width=0.99\columnwidth]{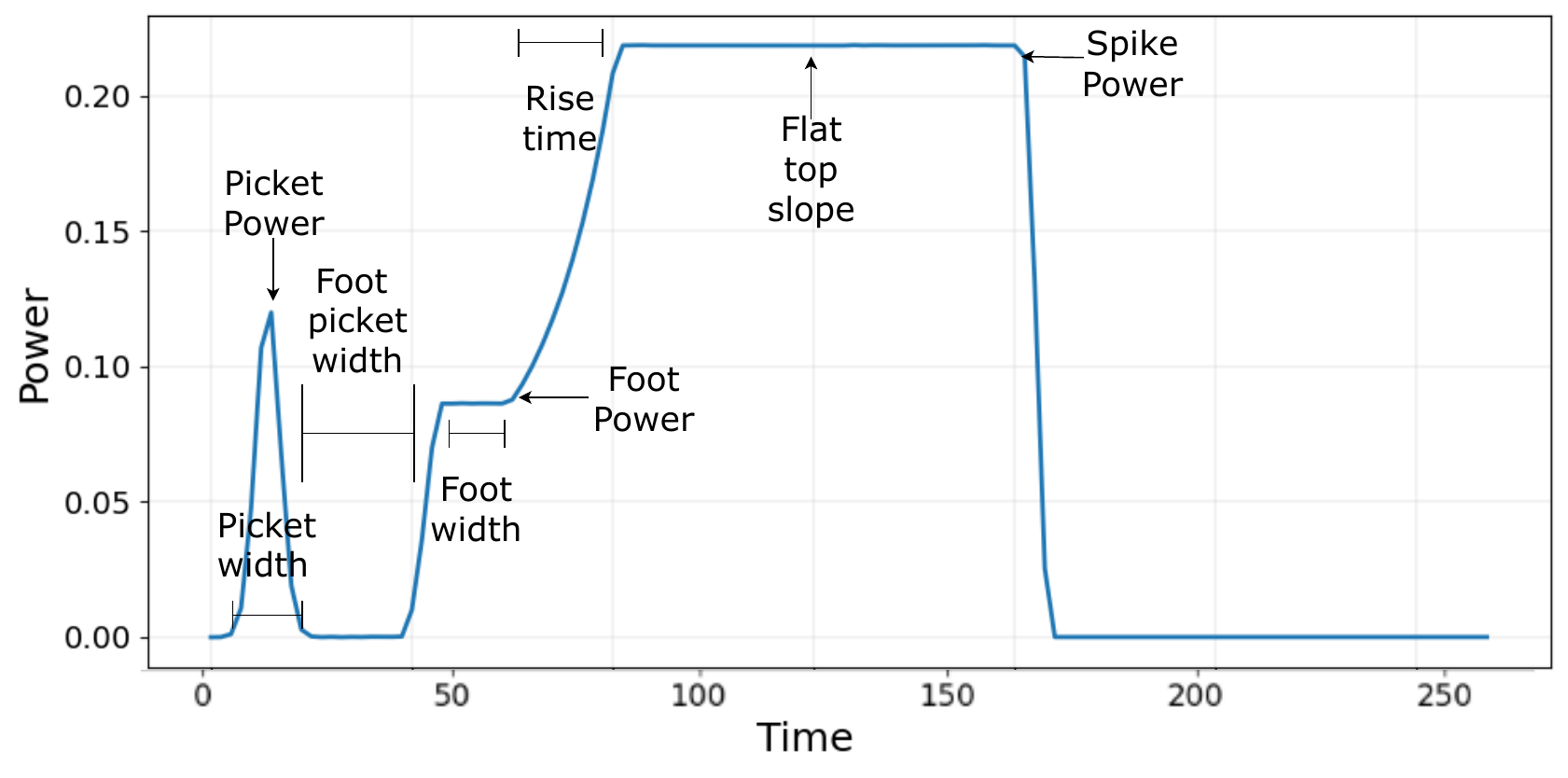}
  \caption{The LP is typically characterized by 12 parameters, some of which are annotated here.}
  \label{fig:sample-pulse}
  \vskip -0.2cm
\end{figure}

Total computation, including training, testing and ablations over all training runs and testing seeds, was conducted using a node with four H100 GPUs, running for approximately 400 hours at an average utilization of 75\%. During training, we executed multiple runs in parallel where GPU and CPU capacity allowed. On a single GPU, one training run takes 13 hours for the diffusion model and 7 hours for the LSTM. The ICF dataset comprised 1 million simulations, with a $70$–$30\%$ train–test split used for training and evaluation across all approaches. We created the training and test sets using scikit-learn's \texttt{train\_test\_split} function. 

The hyperparameters used in our objective $\mathcal{O}(\theta)$ are shown in Table \ref{table:inverse_hyperparameters}. For the learning-rate schedule, we use PyTorch's cosine-annealing scheduler. We use Adam as the optimizer. We use $\tau=0.1$ for the implosion outcome optimization experiments in Section \ref{section:optimization}. 
We used a linear noise scheduler to train the diffusion model. For the forward process we set $N=100$ during training and $N=25$ during generation.

Hyperparameters for the diffusion denoising network and the LSTM are listed in Tables \ref{table:hyperparameters} and \ref{table:lstm_hyperparameters}, respectively. The hyperparameters were selected using the optuna library \citep{optuna_2019} using the following ranges: Embedding layers: $[2:4]$, Hidden dimension: $[512:2048]$, Dropout: $[0.1:0.5]$, Learning rate: $[1e^{-6},1e^{-2}]$, Batch size: $\{64, 128, 256, 512\}$ and when applicable Transformer layers $[2:4]$,  Attention heads: $[4:8]$, Encoder layers $[1:4]$, Decoding layers $[1:4]$.

\begin{table}[h!]
      \centering
      \small
      \begin{tabular}{|ll|}
        \hline
        \multicolumn{1}{|l|}{Transformer layers}    & 4       \\ \hline
        \multicolumn{1}{|l|}{Attention heads}       & 4       \\ \hline
        \multicolumn{1}{|l|}{Feed forward dimension}& 2048     \\ \hline
        \multicolumn{1}{|l|}{Embedding layers}      & 2       \\ \hline
        \multicolumn{1}{|l|}{Embedding dimension} & 1024    \\ \hline
        \multicolumn{1}{|l|}{Dropout}               & 0.1     \\ \hline
        \multicolumn{1}{|l|}{Learning rate}         & $5e^{-5}$ \\ \hline
        \multicolumn{1}{|l|}{Batch size}         & 128 \\ \hline
      \end{tabular}
      \caption{Network Hyperparameters $\epsilon_\theta$ used in diffusion model.}
      \label{table:hyperparameters}
\end{table}

\begin{table}[h!]
      \centering
      \small
      \begin{tabular}{|ll|}
        \hline
        \multicolumn{1}{|l|}{Num of layers}    & 4       \\ \hline
        \multicolumn{1}{|l|}{Number of hidden units} & 512     \\ \hline
        \multicolumn{1}{|l|}{Learning rate}         & $1e^{-5}$ \\ \hline
        \multicolumn{1}{|l|}{Batch size}         & 128 \\ \hline
      \end{tabular}
      \caption{Network Hyperparameters used in the LSTM autoregressive model.}
      \label{table:lstm_hyperparameters}
\end{table}

\begin{table}[h!]
      \centering
      \small
      \begin{tabular}{|ll|}
        \hline
        \multicolumn{1}{|l|}{$\lambda_{\mathcal{S}}$}       & 0.5      \\ \hline
        \multicolumn{1}{|l|}{$\lambda_{\mathcal{P}}$}      & 0.25       \\ \hline
        \multicolumn{1}{|l|}{$\lambda_{\mathcal{L}}$} & 0.25     \\ \hline
      \end{tabular}
      \caption{Hyperparameters used for our inverse-model objective $\mathcal{O}(\theta)$.}
      \label{table:inverse_hyperparameters}
\end{table}


\subsection{LILAC surrogate}
\label{section:LILAC surrogate}
The LILAC surrogate ($\mathbf{S_\phi}$) is a MLP with the following hyper-params. It is trained with the Adam optimizer \cite{kingma2014adam}, with a batch size of 128, learning rate of $1e^{-5}$ for 100 epochs.

\begin{table}[h!]
      \centering
      \small
      \begin{tabular}{|ll|}
        \hline
        \multicolumn{1}{|l|}{Num. of layers}       & 4      \\ \hline
        \multicolumn{1}{|l|}{Num. of hidden units}      & 256       \\ \hline
        \multicolumn{1}{|l|}{Activation function} & ReLU     \\ \hline
      \end{tabular}
      \caption{Hyperparameters used for the LILAC surrogate $\mathbf{S_\phi}$}
      \label{table:lilac_hyperparameters}
\end{table}

The surrogate has an average error of 1.4\% across all the outputs.

\paragraph{Choice of surrogate architecture.}
The surrogate solves a fixed-size regression from $(\mathbf{l}, \mathbf{p})$ to the scalar outcome vector $\mathbf{m}$, rather than a function-to-function mapping on a space-time grid. We therefore use an MLP, and compare it against two architectures from the scientific machine learning literature: a one-dimensional Fourier neural operator (FNO) \citep{li2021fno, kovachki2023neuraloperator} with 32 Fourier modes and width 64, and a physics-informed MLP \citep{raissi2019pinn} with the same capacity as our MLP, $\tanh$ activations, and a soft energy-conservation residual added to the training loss. Table~\ref{table:surrogate_architectures} reports the test MAPE of each surrogate.

\begin{table}[h!]
      \centering
      \small
      \begin{tabular}{lc}
        \toprule
        \textbf{Surrogate} & \textbf{Test MAPE (\%) $\downarrow$} \\
        \midrule
        MLP (used in LPDS) & $\mathbf{1.40}$ \\
        FNO (1D, 32 modes, width 64) & 1.43 \\
        Physics-informed MLP ($\tanh$, energy-conservation residual) & 2.31 \\
        \bottomrule
      \end{tabular}
      \caption{Comparison of surrogate architectures for approximating the LILAC outcomes $\mathbf{m}$.}
      \label{table:surrogate_architectures}
\end{table}

The FNO matches but does not improve on the MLP, as its spectral basis offers no advantage when the target is a vector of scalar summaries of the LILAC solution. The physics-informed variant underperforms for three reasons: the outcomes $\mathbf{m}$ are summaries of the full LILAC solution rather than the spatio-temporal fields on which a PDE residual is naturally defined; the simulated training data already satisfy energy conservation, so the additional soft penalty is redundant and biases the fit; and $\tanh$ activations underperform ReLU on this regression. A neural-operator surrogate that predicts the full LILAC fields is a possible future direction, but would require a substantially larger output space without changing the outcome error that LPDS optimizes against.

\subsection{Mapping Function}
\label{section:mapping_function}
The mapping function ($\mathcal{M}$) is a 1 layer bidirectional LSTM with 64 hidden units. We concatenate the hidden state from both directions, before passing it through an output linear layer. It is trained with the Adam optimizer, with a batch size of 64, learning rate of $1e^{-5}$ for 250 epochs. The model achieves an average error of 1.1\% across all 12 parameters in $C$.

\subsection{Data Generation}
Inputs for experiments and LILAC simulations are specified by target and the laser pulse shape. The laser pulse shape is prescribed using an internal
library that parameterizes the laser pulse shape into an ad-hoc descriptive
featurization such as the time delay between gaussian shaped features,
maximum power and total energy. Therefore, the input is defined within a 20-dimensional parameter space, consisting of target specifications and a parameterized laser-pulse description. Although this space can be uniformly sampled under physical constraints (e.g., maximum laser energy, allowable target thicknesses), random sampling of the 20-D hypervolume does not represent the way experimental and consequently simulation designs are chosen. In practice, high-performing implosions occupy only a small subset of this high-dimensional space, guided by physics intent, rather than being distributed throughout the full admissible hypervolume.
Consequently, the region of desirable inputs does not fill the full 20-D space but instead lies on a lower-dimensional manifold embedded within it. This manifold is not explicitly known or parameterized: it emerges from complex physics couplings between target geometry and laser-pulse features. To more effectively sample this region, we supplement random draws with perturbations of known high-performing implosions. Perturbing their laser-pulse features and target parameters produces a family of nearby points that remain close to the underlying manifold while still providing variation in implosion performance. This strategy implicitly captures the trait that for a fixed target specification, laser-pulse shapes close to physics-based designs typically outperform large random perturbed ones. The resulting dataset therefore reflects a distribution with multiple high-performance pockets concentrated on an unknown manifold within the 20-D input space, rather than a uniform distribution over the entire hypervolume.


\section{Constrained Optimization}
\label{appendix:constrained_optimization}
We also try an approach that completely forgoes training a predictive model to estimate the pulse shape by formulating the inverse modeling problem as a constrained optimization problem. The laser pulse $\mathbf{l}$ can be represented as a combination of 12 real-valued parameters $\{c_i\}_{i=1}^N \subseteq \mathbf{c}$ that completely define the shape of the pulse. As described in \ref{section:auxiliary_objective}, we now train a surrogate $S_\phi$ that maps $\mathbf{c}$ and the target pellet parameters $\mathbf{p}$ to the LILAC outcomes $\mathbf{m'}$. The inverse problem can now be formulated as a constrained optimization problem over the parameters $\{c_i\}_{i=1}^{12}$ that minimizes the following mean squared error.

\begin{equation}
\mathcal{L}_{\text{surrogate}}(\phi) = \mathbb{E}_{(\mathbf{c}, \mathbf{m}, \mathbf{p}) \sim D_F} \left[ \left\| \mathbf{m} - \mathbf{S}_\phi(\mathbf{c}, \mathbf{p}) \right\|^2 \right].
\end{equation}

Note that since $S_\phi$ is differentiable, we can also use gradient-based optimization techniques. The pitfall of this approach is that it often produces pulses that are out-of-distribution for the surrogate $S_\phi$, leading to predictions $\mathbf{m'}$ that are unreliable. Also, the optimization routine needs to be run for every sample. With this technique, we obtain an implosion-outcome error of 10\%, compared with $1.93\%$ for $\mathbf{LPDS}_\text{Diff}$ under the same surrogate (Table~\ref{table:experimental_results}). The gap reflects the limits of optimizing a 12-dimensional parameterization: these parameters compress away fine temporal structure of the 256-dimensional pulse that the surrogate uses to match the target outcomes, whereas LPDS generates the full pulse directly.

\section{Comparison with Off-the-Shelf Conditional Generators}
\label{appendix:offtheshelf_baselines}

LPDS performs amortized conditional posterior estimation of $p(\mathbf{l}\mid\mathbf{m},\mathbf{p})$, in the spirit of neural posterior estimation in simulation-based inference \citep{cranmer2020frontier}, in a setting where the simulator is too expensive for online likelihood evaluation. To position LPDS against existing approaches, we compare it with seven off-the-shelf methods that cover the main families of conditional generation and inverse-problem solvers: a variational autoencoder (VAE) \citep{Kingma2013}, a generative adversarial network (GAN) \citep{goodfellow2014gan}, tabular diffusion (TabSyn) \citep{zhang2024mixedtype}, variational latent diffusion for inverse problems in high-energy physics (VLD) \citep{shmakov2023endtoend}, a conditional invertible neural network (cINN) \citep{ardizzone2019analyzing}, test-time optimization in the latent space of a generative model against the surrogate (CSGM) \citep{bora2017csgm}, and a deterministic neural-operator inverse map (FNO) \citep{li2021fno, kovachki2023neuraloperator}. All methods use the same conditioning inputs $(\mathbf{m}, \mathbf{p})$, the same training set $D_F$, the same test set, and the same metrics as Table~\ref{table:experimental_results}. For VLD, we use the end-to-end architecture but not its consistency loss, which does not apply to the ICF setting. The cINN uses 12 Glow-style affine coupling blocks with 512 hidden units and is trained by maximum likelihood only, without physics feedback. CSGM optimizes the latent variable of the conditional VAE at test time against the frozen surrogate $\mathbf{S}_\phi$. The FNO baseline is a one-dimensional Fourier neural operator trained to map $(\mathbf{m}, \mathbf{p})$ directly to the pulse.

\begin{table*}[h]
\centering
\resizebox{\textwidth}{!}{%
\begin{tabular}{lcccc}
\toprule
\textbf{LPDS variant / Baseline} & \textbf{Diversity} $\uparrow$ & $\mathbf{m}$\textbf{ Error} $\downarrow$ & \textbf{Reconstruction} $\downarrow$ & \textbf{Energy Cons.} $\downarrow$ \\
 & & (\%) & ($\times 10^{-3}$) & (\%) \\
\midrule
LPDS Diff & $0.64\pm0.077$ & $1.93\pm0.046$ & $5.6\pm0.20$ & $1.59\pm0.14$ \\
LPDS LSTM & -- & $1.65\pm0.09$ & $0.1\pm0.05$ & $0.66\pm0.07$ \\
\midrule
VAE    & $0.16\pm0.06$ & $4.97\pm0.24$  & $1.0\pm0.10$   & $0.61\pm0.04$ \\
GAN    & $0.68\pm0.09$ & $9.88\pm0.73$  & $4.0\pm0.20$   & $0.78\pm0.09$ \\
TabSyn & $0.66\pm0.13$ & $15.80\pm0.80$ & $23.0\pm0.20$  & $4.76\pm0.07$ \\
VLD    & $0.38\pm0.06$ & $16.60\pm0.96$ & $16.0\pm0.06$  & $1.92\pm0.08$ \\
cINN   & $0.72\pm0.11$ & $9.91\pm0.01$  & $14.75\pm0.40$ & $1.27\pm0.03$ \\
CSGM   & $0.01\pm0.01$ & $2.58\pm0.00$  & $0.56\pm0.50$  & $0.75\pm0.02$ \\
FNO (inverse) & -- & $5.04\pm0.01$ & $0.52\pm0.48$ & $0.70\pm0.01$ \\
\bottomrule
\end{tabular}
}
\caption{Comparison of LPDS with off-the-shelf conditional generators and inverse-problem methods on the simulation test set. $\pm$ denotes standard deviation over seeds and training runs. Diversity is not defined for deterministic methods.}
\label{table:offtheshelf_baselines}
\end{table*}

Table~\ref{table:offtheshelf_baselines} highlights two failure modes that LPDS is designed to avoid. Amortized conditional generators trained without physics feedback (VAE, GAN, TabSyn, VLD, cINN, FNO) leave a large gap in outcome consistency, with $\mathbf{m}$ errors between $4.97\%$ and $16.60\%$, compared with $1.93\%$ for $\mathbf{LPDS}_\text{Diff}$. In particular, the cINN, a flow-based approach of the kind used for neural posterior estimation, reaches high diversity but a $9.91\%$ $\mathbf{m}$ error, illustrating the difficulty of scaling tractable-density posteriors to a 256-dimensional structured output. Conversely, test-time optimization against the surrogate (CSGM) attains a competitive $\mathbf{m}$ error of $2.58\%$ only by collapsing diversity to $0.01$, and must be rerun for every new $(\mathbf{m}, \mathbf{p})$. LPDS combines an expressive conditional generator with surrogate and physics feedback during training ($\mathcal{S}$, $\mathcal{P}$), and is the only approach that combines an $\mathbf{m}$ error below $2\%$ with high diversity, while keeping energy violation below $2\%$.

\begin{figure}[h!]
    \centering
    \begin{subfigure}[b]{0.49\textwidth}
        \centering
        \includegraphics[width=\textwidth]{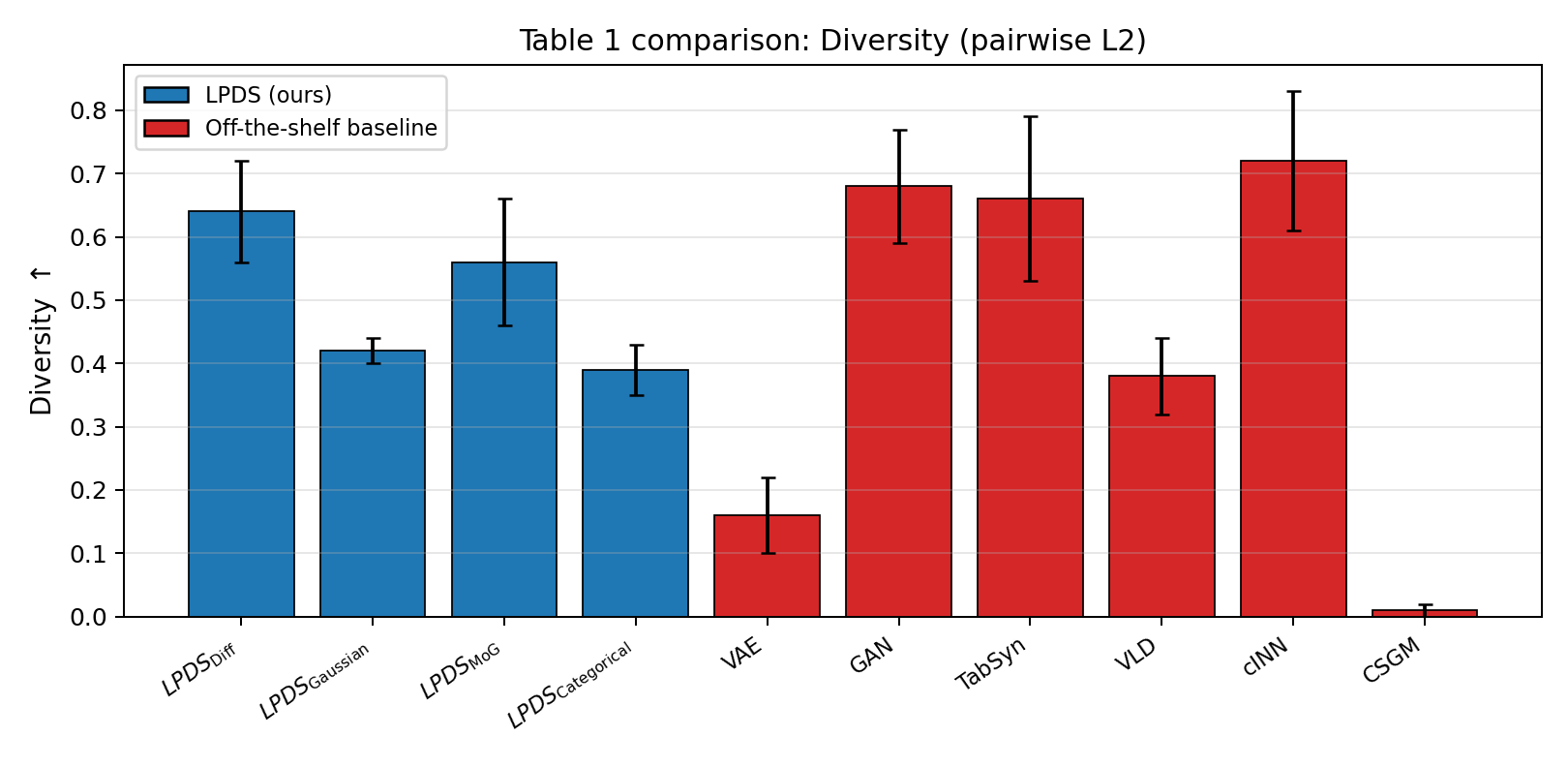}
        \caption{Diversity ($\uparrow$).}
    \end{subfigure}
    \hfill
    \begin{subfigure}[b]{0.49\textwidth}
        \centering
        \includegraphics[width=\textwidth]{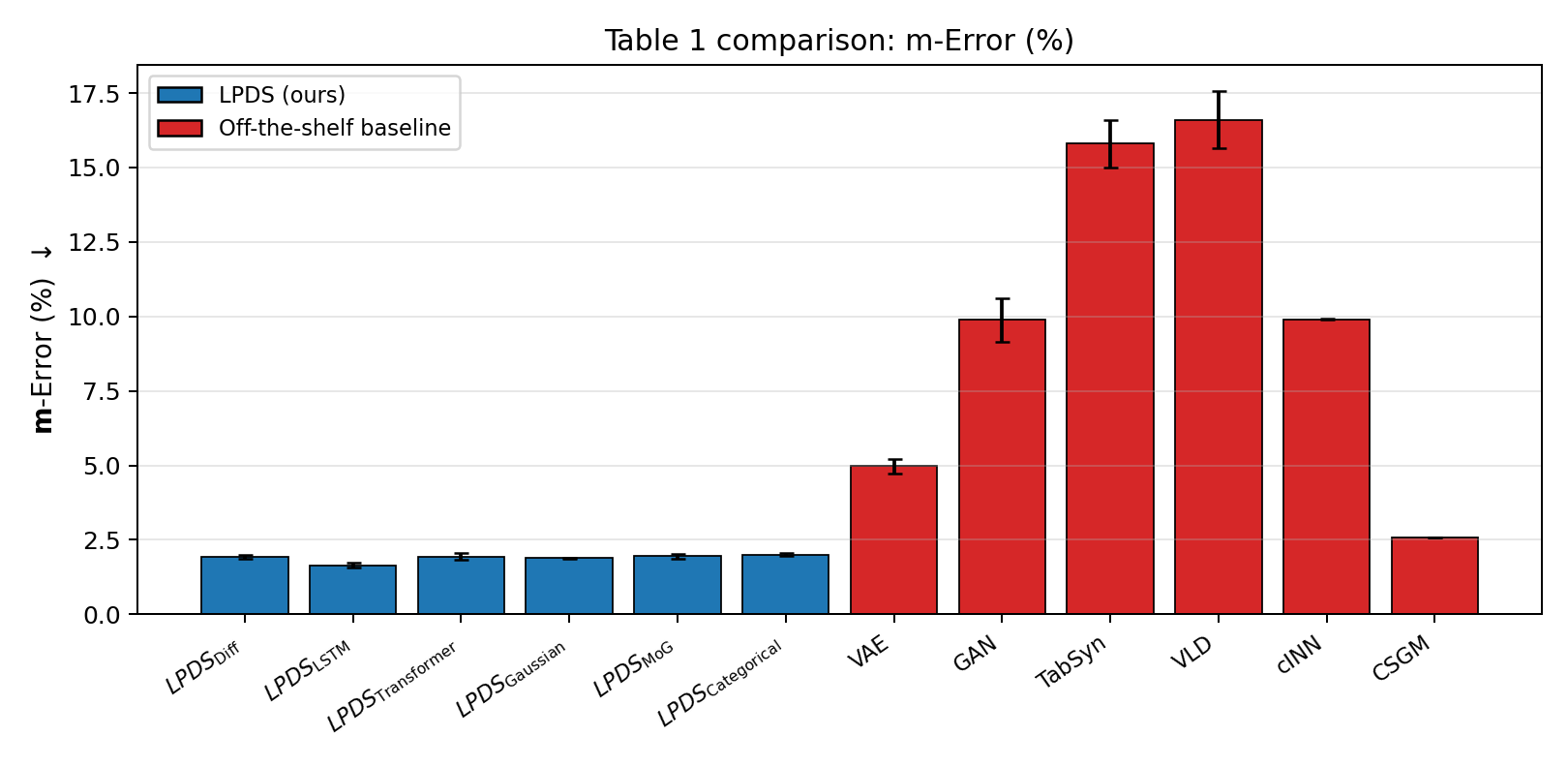}
        \caption{$\mathbf{m}$ error (\%, $\downarrow$).}
    \end{subfigure}
    \begin{subfigure}[b]{0.49\textwidth}
        \centering
        \includegraphics[width=\textwidth]{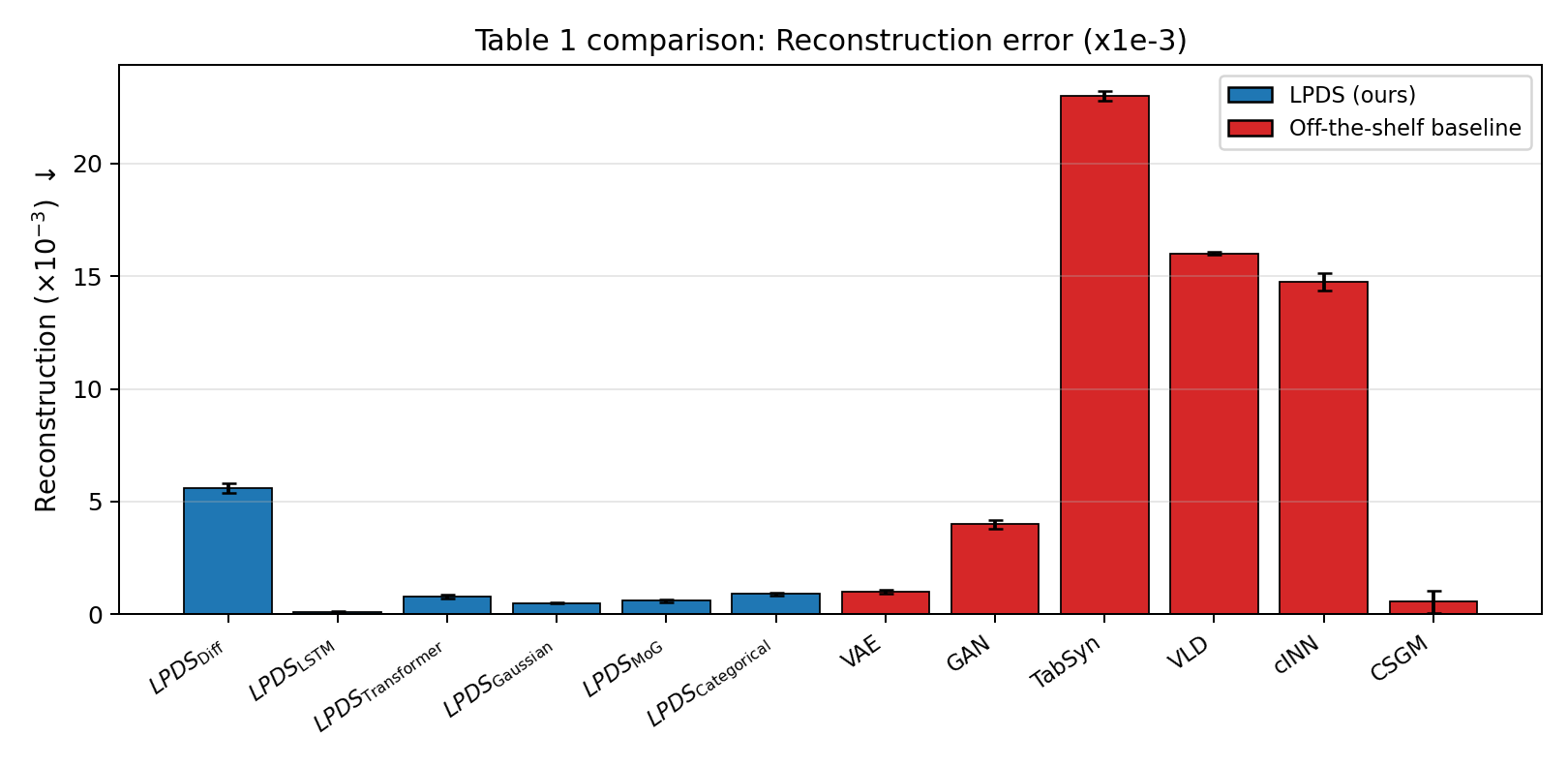}
        \caption{Reconstruction error ($\downarrow$).}
    \end{subfigure}
    \hfill
    \begin{subfigure}[b]{0.49\textwidth}
        \centering
        \includegraphics[width=\textwidth]{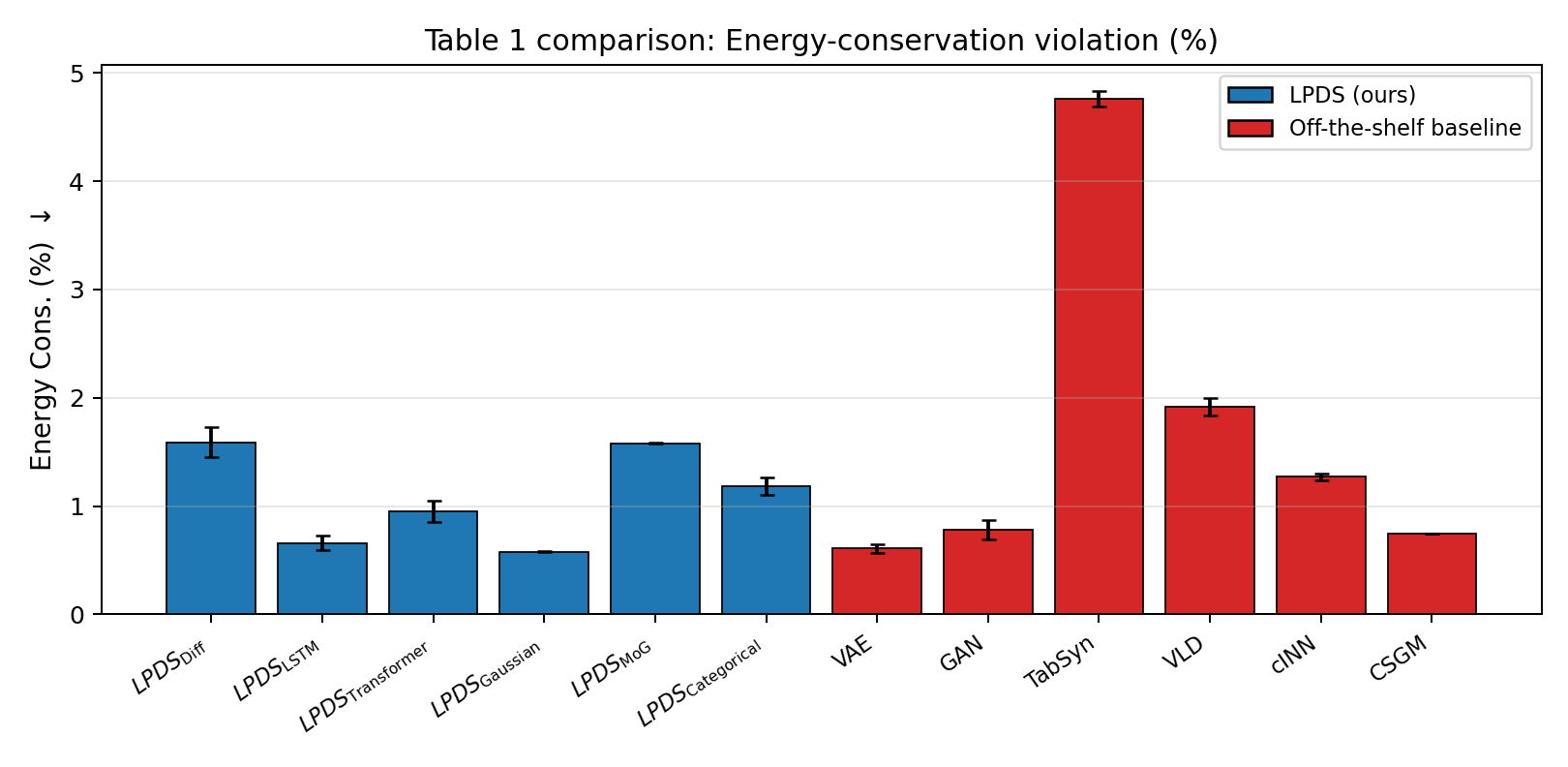}
        \caption{Energy-conservation violation (\%, $\downarrow$).}
    \end{subfigure}
    \caption{LPDS variants compared with off-the-shelf baselines on each metric. Error bars denote standard deviation over seeds and training runs. Diversity is omitted for deterministic models.}
    \label{fig:bar_main}
\end{figure}

\begin{figure}[h!]
    \centering
    \begin{subfigure}[b]{0.49\textwidth}
        \centering
        \includegraphics[width=\textwidth]{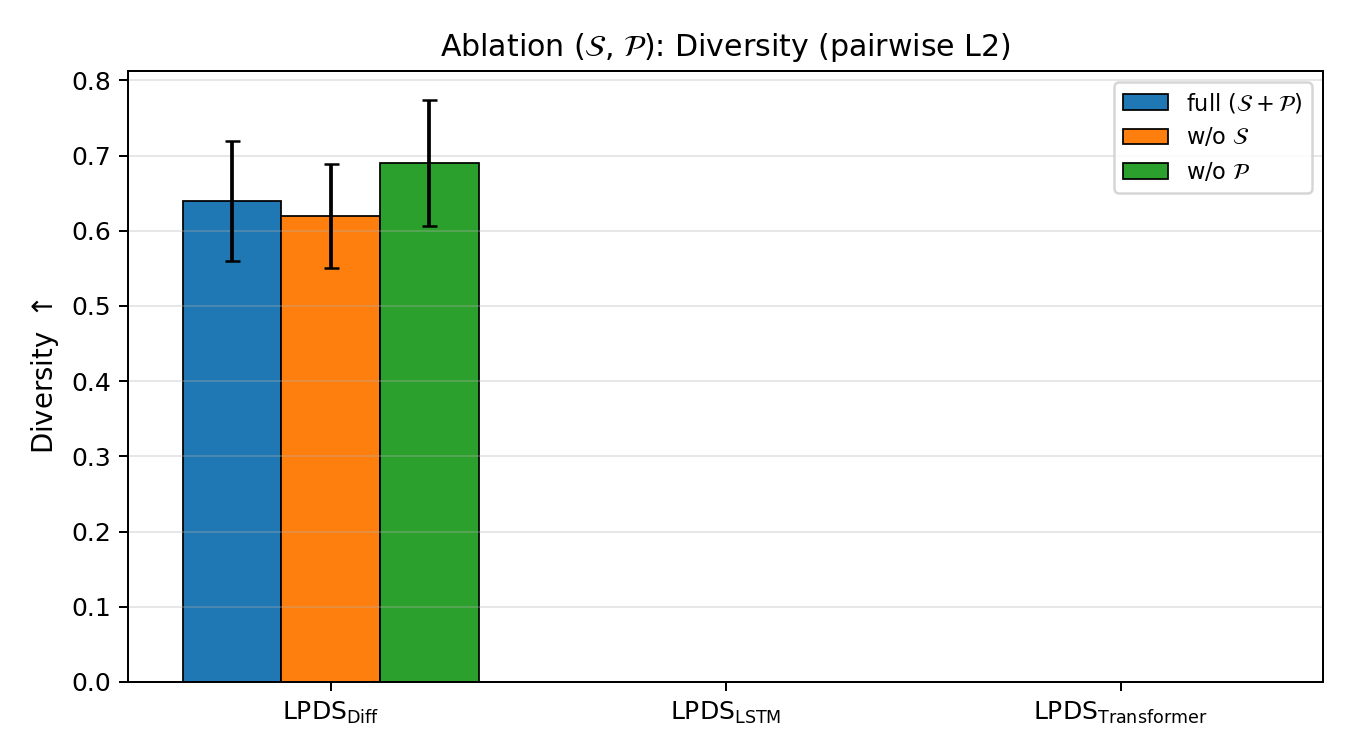}
        \caption{Diversity ($\uparrow$).}
    \end{subfigure}
    \hfill
    \begin{subfigure}[b]{0.49\textwidth}
        \centering
        \includegraphics[width=\textwidth]{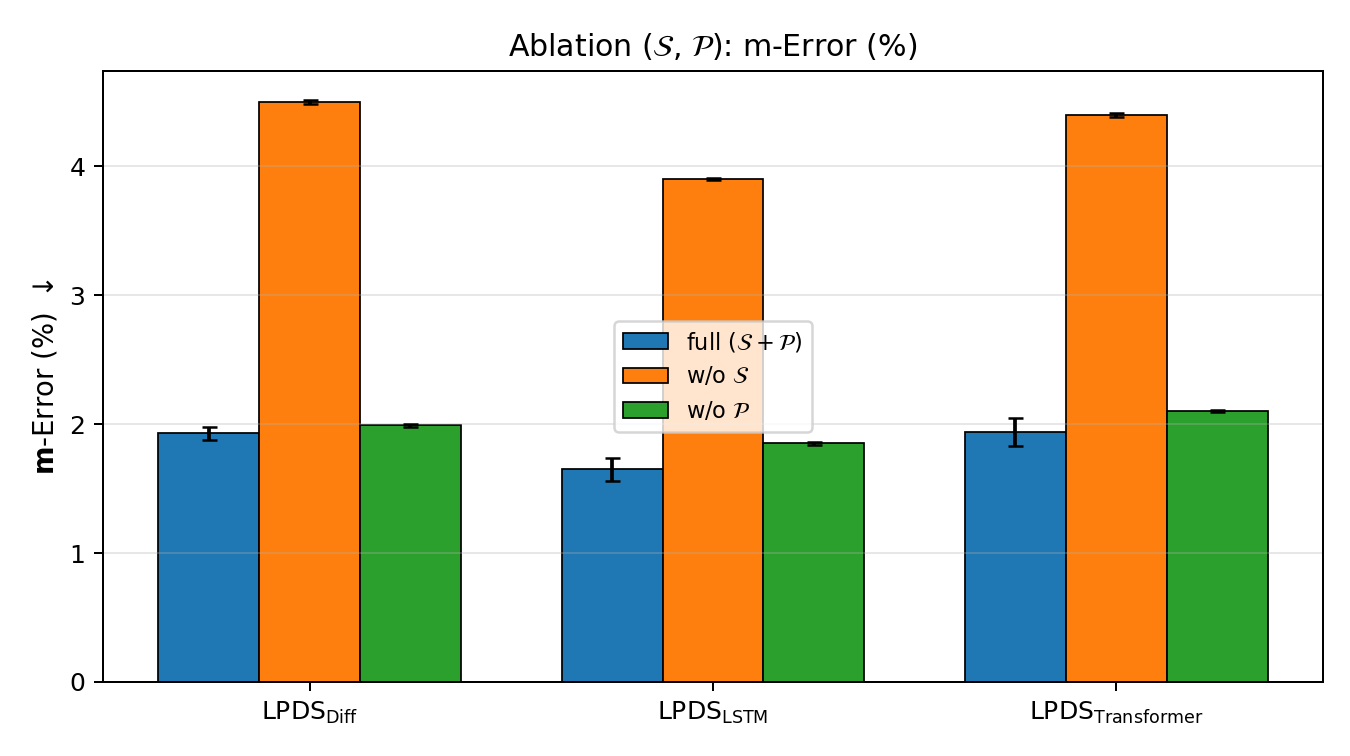}
        \caption{$\mathbf{m}$ error (\%, $\downarrow$).}
    \end{subfigure}
    \begin{subfigure}[b]{0.49\textwidth}
        \centering
        \includegraphics[width=\textwidth]{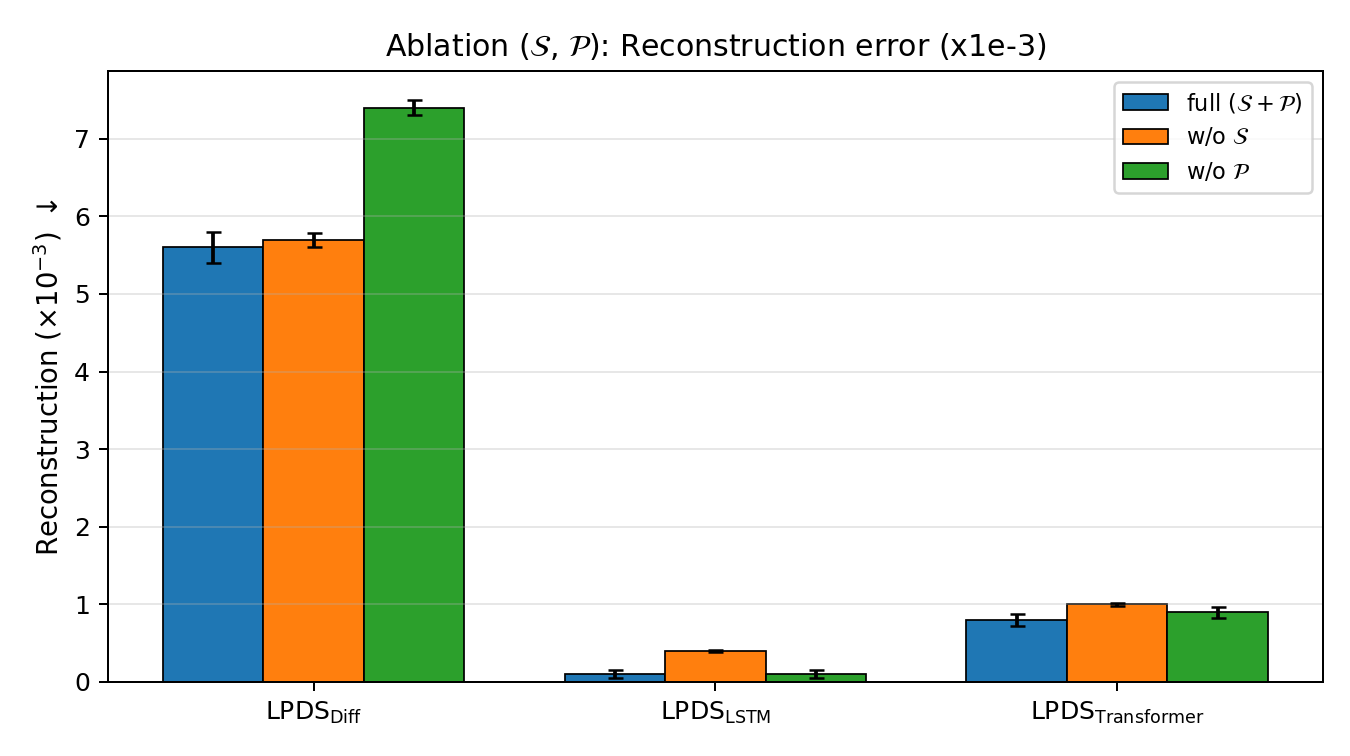}
        \caption{Reconstruction error ($\downarrow$).}
    \end{subfigure}
    \hfill
    \begin{subfigure}[b]{0.49\textwidth}
        \centering
        \includegraphics[width=\textwidth]{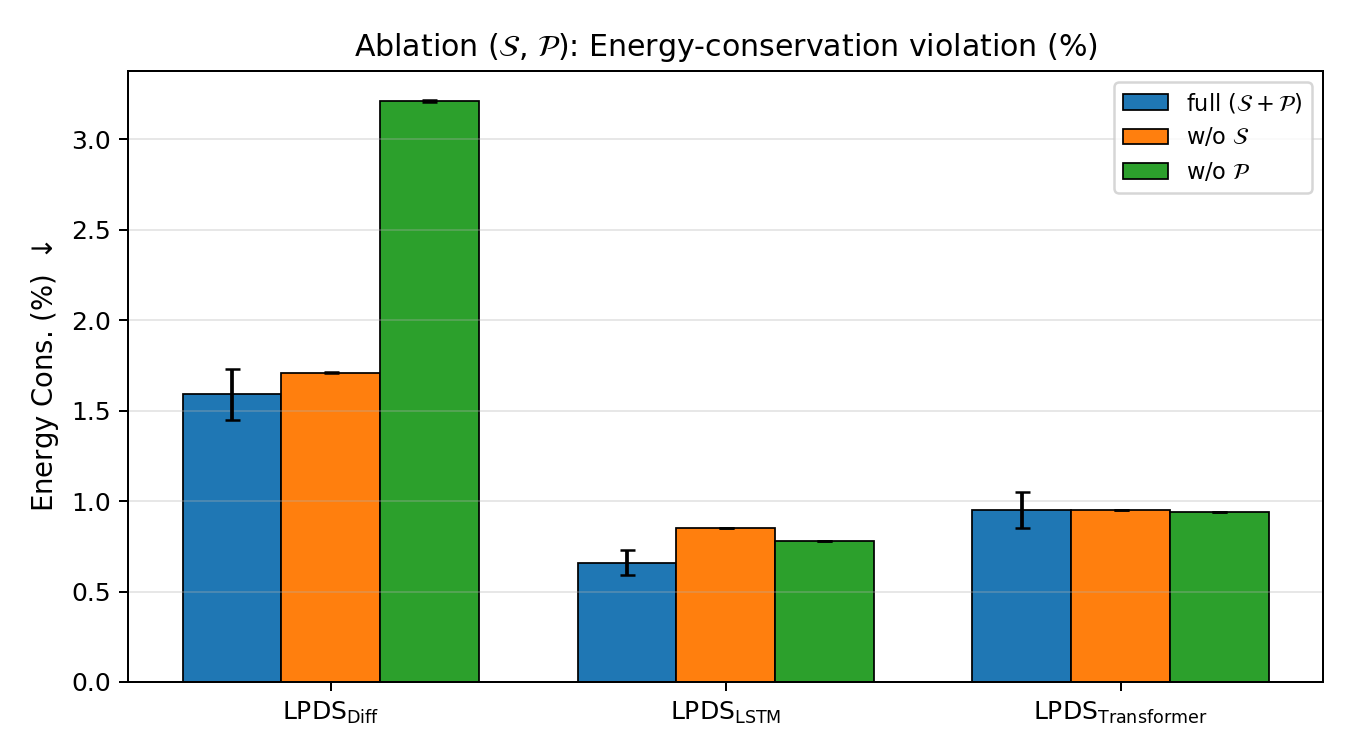}
        \caption{Energy-conservation violation (\%, $\downarrow$).}
    \end{subfigure}
    \caption{Ablation of the surrogate loss $\mathcal{S}$ and the physics-informed loss $\mathcal{P}$ (Table~\ref{table:experimental_results}). Removing $\mathcal{S}$ more than doubles the $\mathbf{m}$ error of all three models, while removing $\mathcal{P}$ roughly doubles the energy violation of $\mathbf{LPDS}_\text{Diff}$.}
    \label{fig:bar_ablation}
\end{figure}

Figure~\ref{fig:bar_main} summarizes Tables~\ref{table:experimental_results} and~\ref{table:offtheshelf_baselines} per metric, and Figure~\ref{fig:bar_ablation} visualizes the effect of removing $\mathcal{S}$ and $\mathcal{P}$.

\section{Ablation Studies}
\subsection{Trade-off between Fidelity and Diversity}
\label{sec:fid-div}
We performed an ablation study by systematically varying the relative weights assigned to the reconstruction term ($\lambda_{\mathcal{L}}$) and the outcome-consistency term ($\lambda_{\mathcal{S}}$) in the composite loss \ref{eq:composite_equation}, while keeping the physics penalty weight ($\lambda_{\mathcal{P}}$) constant. This analysis isolates the inherent trade-off surface between generating highly faithful, but potentially less diverse, laser pulse shapes (high $\lambda_{\mathcal{L}}$) and generating diverse solutions that effectively satisfy the target outcome (high $\lambda_{\mathcal{S}}$).

\begin{figure}[tbp]
    \centering
    
    \begin{subfigure}[l]{0.4\textwidth}
        \centering
        \includegraphics[width=\textwidth]{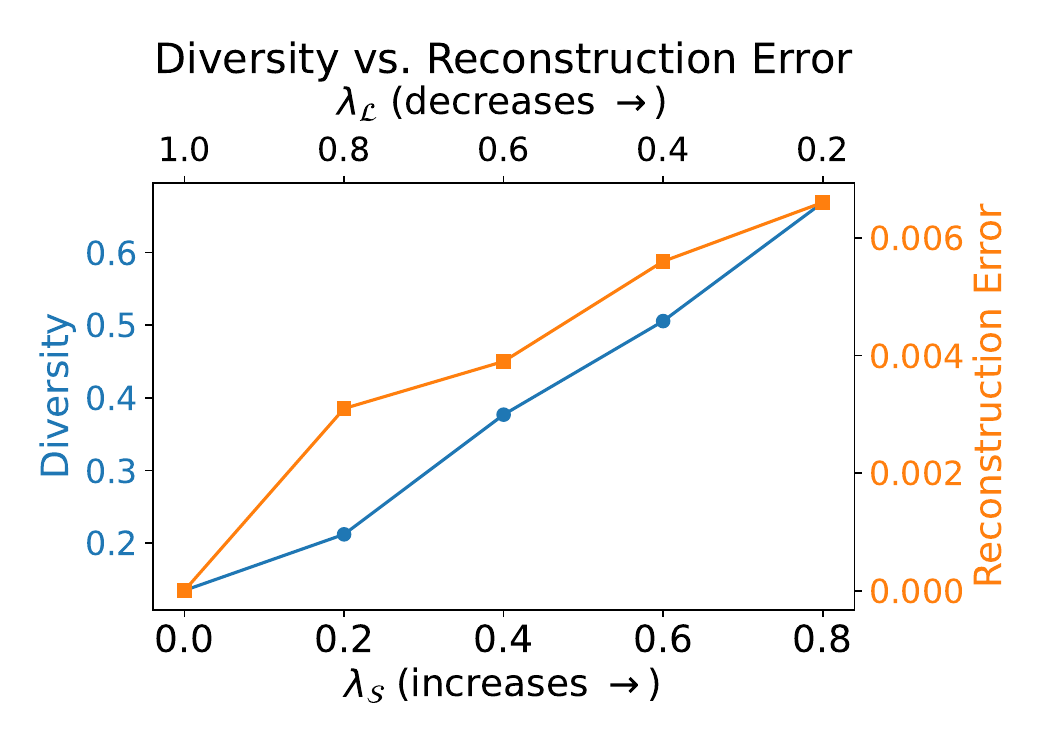}
        \caption{Fidelity–Diversity trade-off curve.}
        \label{fig:div-fid}
        
    \end{subfigure}
    \begin{subfigure}[r]{0.48\textwidth}
        \centering
        \begin{tabular}{cccc}
            \toprule
            $\lambda_{\mathcal{L}}$ & $\lambda_{\mathcal{S}}$ & $\mathcal{L}$ & Diversity \\
            \midrule
            1.0 & 0.0 & $1.0\times10^{-5}$ & 0.130\\
            0.8 & 0.2 & 0.0031 & 0.21 \\
            0.6 & 0.4 & 0.0039 & 0.37 \\
            0.4 & 0.6 & 0.0056 & 0.50 \\
            0.2 & 0.8 & 0.0064 & 0.66 \\
            \bottomrule
        \end{tabular}
        \caption{Metrics for different hyperparameters.}
        \label{tab:metrics}
    \end{subfigure}
    
    \caption{Fidelity–Diversity analysis: (a) trade-off curve and (b) measured metrics across $\lambda_{\mathcal{L}}$ and $\lambda_{\mathcal{S}}$ settings.}
    \label{fig:combined}
\end{figure}

Figure \ref{fig:combined} illustrates this critical trade-off. As the model's objective shifts towards maximizing exploration of the design space (increasing $\lambda_{\mathcal{S}}$), it must tolerate a corresponding degradation in the reconstruction fidelity ($\mathcal{L}$).

The results demonstrate a clear Pareto frontier. When $\lambda_{\mathcal{L}}=1.0$ and $\lambda_{\mathcal{S}}=0.0$ (strong emphasis on reconstruction), the model acts primarily as an autoencoder, achieving near-perfect fidelity ($\mathcal{L}=1.0 \times 10^{-5}$) but minimal design diversity ($0.13$). Conversely, as the weight shifts to strongly prioritize the surrogate outcome ($\lambda_{\mathcal{S}}=0.8$), diversity reaches its maximum ($0.66$) at the expense of fidelity ($\mathcal{L}=0.0064$). We excluded $\lambda_{\mathcal{L}}=0.0$ and $\lambda_{\mathcal{S}}=1.0$ because the model failed to learn physically viable representations.

\subsection{Physical Violations}
This section evaluates the violation rates for the two primary operational constraints investigated: the instantaneous per-timestep maximum power caps and the maximum cumulative energy allowed during ICF experiments. These rates, detailed in Table \ref{tab:violation_rates}, are crucial indicators of the robustness and feasibility of the generated LPs.

\begin{table}
\centering
\begin{tabular}{lcc}
\toprule
\textbf{Approach} & Maximum Power Cap & Maximum Cumulative Energy\\
\midrule
LSTM & 0$\%$ & 0$\%$ \\
Transformer & 0$\%$ & 0$\%$ \\
Diffusion & 0$\%$ & 0$\%$ \\
LSTM$_{\text{Gaussian}}$ & 0$\%$ & 0$\%$  \\
LSTM$_{\text{MixtureOfGaussian}}$ & 0$\%$ & 0$\%$\\
LSTM$_{\text{Categorical}}$ & 0$\%$ & 0$\%$  \\

\bottomrule

\end{tabular}
\caption{Violation rates of the benchmarked approaches for LP generation.}
\label{tab:violation_rates}
\end{table}

The data in Table \ref{tab:violation_rates} show that all approaches, including all LSTM variants (standard, Gaussian, mixture of Gaussians, and categorical), the Transformer, and the diffusion model, achieved 0\% violation rates for both the instantaneous power cap and the cumulative energy constraint on this evaluation. This outcome indicates that $\mathcal{L}$ and $\mathcal{P}$ provide sufficient guidance during training for generation to remain within physically viable designs in the evaluated setting, without explicit constraint enforcement at inference time.

\subsection{Generalization to Unseen Pellet Configurations}
\label{sec:pellet_stratified}

The random 70\%--30\% split used in the main experiments may place pulses from the same perturbation family in both the training and test sets. To evaluate generalization to genuinely unseen target designs, we construct a pellet-stratified split: we cluster the dataset with $k$-means ($k=10$) on the pellet parameters $\mathbf{p}$ and hold out the two clusters closest to the median cluster size, so that test pellet configurations are disjoint from those seen during training. Table~\ref{tab:pellet_stratified} reports the $\mathbf{m}$ error of $\mathbf{LPDS}_\text{Diff}$ on the held-out clusters. As a reference, we also report the $\mathbf{m}$ error obtained by passing the ground-truth pulses of the held-out clusters through the surrogate, which gives the error floor attainable on these clusters.

\begin{table}[h!]
\centering
\begin{tabular}{lc}
\toprule
\textbf{Setting} & $\mathbf{m}$\textbf{ Error (\%) $\downarrow$} \\
\midrule
Random split, $\mathbf{LPDS}_\text{Diff}$ (Table~\ref{table:experimental_results}) & 1.93 \\
Pellet-stratified split, ground-truth pulses (surrogate floor) & 1.53 \\
Pellet-stratified split, $\mathbf{LPDS}_\text{Diff}$ & 2.02 \\
\bottomrule
\end{tabular}
\caption{Generalization of $\mathbf{LPDS}_\text{Diff}$ to held-out pellet clusters.}
\label{tab:pellet_stratified}
\end{table}

On the held-out pellet clusters, $\mathbf{LPDS}_\text{Diff}$ reaches a $2.02\%$ $\mathbf{m}$ error, only $0.49$ percentage points above the surrogate floor and $0.09$ percentage points above its performance on the random split. This indicates that LPDS generalizes to unseen pellet configurations rather than interpolating within known perturbation families.

\subsection{Computational Efficiency}
\label{sec:computational_efficiency}

We analyze the computational overhead for training and inference (sampling) across the different approaches. While the training and inference costs are negligible when considering the month lead time and scale of the ICF experimental cycle, we study these properties because they are paramount for evaluating the model's scalability and applicability to other scientific inverse design tasks where time-to-solution is a critical constraint. All reported times are based on the execution using a single NVIDIA H100 GPU for both training and inference tasks.

Table~\ref{tab:computational_benchmarks} presents the average training duration, samples generated per second and latency in milliseconds (time per batch). The training time is an average over 5 independent runs, with convergence defined as the point where the validation loss did not meaningfully improve over 20 consecutive epochs. The sampling time reports the average samples/time,  when evaluated over the entire test set. These results highlight the intrinsic trade-offs among model complexity, training investment, and broader scientific utility.

\begin{table}[h!]
\centering
\begin{tabular}{lccc}
\toprule
\textbf{Approach} & \textbf{Avg. Training Time (h)} & \textbf{Throughput (sample/s)} & Latency (ms)\\
\midrule
LSTM & 10.21$\pm$0.23 & 16666 & 26.8 \\
Transformer & 8.42$\pm$0.45 & 34666 & 18.3 \\
Diffusion & 29.58$\pm$1.4 & 1939 & 528\\
LSTM$_{\text{Gaussian}}$ & 12.21$\pm$0.66 & 16666 & 26.8 \\
LSTM$_{\text{MixtureOfGaussian}}$ & 13.43$\pm$0.75 & 16666 & 26.8\\
LSTM$_{\text{Categorical}}$ & 12.13$\pm$0.8 & 16666 & 26.8 \\

\bottomrule

\end{tabular}
\caption{Computational efficiency of the different approaches used for ICF LP generation.}
\label{tab:computational_benchmarks}
\end{table}

\subsection{Practical Guidance for Inverse Design in ICF}
\label{sec:decision_recipe}

The systematic benchmarking and ablation studies provide concrete, data-driven guidance for selecting the optimal generative strategy based on the specific goals of ICF LP generation. The decision-making process involves two stages: 1) selecting the appropriate generative architecture (Table~\ref{table:experimental_results}), and 2) tuning the composite loss weights ($\lambda_{\mathcal{L}}, \lambda_{\mathcal{S}}$) to manage the Bias-Variance trade-off (Figure~\ref{fig:combined}).

\paragraph{Model Architecture Selection}
The choice of model is primarily dictated by the required balance between solution fidelity and design diversity(Table~\ref{table:experimental_results}):

\begin{itemize}
    \item \textbf{Scenario 1: High Fidelity / Low Risk (Interpolation):} If the goal is to generate LPs closely resembling known, successful designs from the training set (e.g., for marginal improvements or verification), models with low Reconstruction Error and low $\mathbf{m}$ Error are preferred. The Deterministic LSTM network is ideal due to its high fidelity and strong Energy Conservation. These models act as robust interpolators but offer no novelty.
    
    \item \textbf{Scenario 2: Balanced Exploration (Best Generalist):} If the objective is to generate novel, diverse LPs while maintaining strong implosion performance, a probabilistic model is required. The Diffusion Model  and LSTM$_{\text{Gaussian}}$ offer the best trade-off. The Diffusion model offers superior diversity, while the LSTM$_{\text{Gaussian}}$ is more physically consistent and computationally cheaper.
\end{itemize}

\paragraph{Hyperparameter Guidance}
Once the architecture is chosen, the weights $(\lambda_{\mathcal{L}}, \lambda_{\mathcal{S}})$ must be set based on the desired level of novelty versus realism. The ablation study in Figure~\ref{fig:combined} explicitly maps this trade-off:

\begin{itemize}
    \item \textbf{Conservative Setting (Minimize Variance):} To minimize the risk of generating non-physical, low-fidelity LPs, one should set $\lambda_{\mathcal{L}} \gg \lambda_{\mathcal{S}}$. This corresponds to the region of the Pareto curve where $\mathcal{L}$ is lowest (e.g., $\lambda_{\mathcal{L}}=1.0$, $\mathcal{L}=1.0 \times 10^{-5}$). This strategy minimizes the solution variance but introduces strong bias toward the training distribution, limiting novelty.
    
    \item \textbf{Exploratory Setting (Minimize Bias):} To maximize the generation of novel LPs (minimizing bias against unexplored regions), one should accept a higher reconstruction loss. This corresponds to the highest diversity region (e.g., $\lambda_{\mathcal{L}}=0.2$, Div $=0.8$). This strategy maximizes novelty but increases the variance and risk of generating unstable pulses.
    
    \item \textbf{Optimal Balance:} For our application, the optimal choice lies at the middle of the Fidelity-Diversity Pareto frontier (e.g., $\lambda_{\mathcal{L}}=0.4, \lambda_{\mathcal{S}}=0.6$). This setting maximizes the gain in diversity (novelty) for the smallest corresponding cost in reconstruction fidelity (realism), offering the best overall stability and design utility.
\end{itemize}

\section{Theoretical Remarks}
\subsection{Convergence of the composite objective}
Equation \ref{eq:composite_equation} defines the composite loss
\begin{equation}
\mathcal{O}(\theta) = \lambda_{\mathcal{L}} \mathcal{L}(\theta) + \lambda_{\mathcal{S}} \mathcal{S}(\theta) + \lambda_{\mathcal{P}} \mathcal{P}(\theta),
\end{equation}
where $\mathcal{L}(\theta)$ denotes the reconstruction or likelihood term, $\mathcal{S}(\theta)$ the outcome-consistency term using the differentiable surrogate $S_\phi$, and $\mathcal{P}(\theta)$ the physics penalty.
Because $\mathcal{O}(\theta)$ is a nonconvex objective, global optimality cannot be guaranteed. 
However, under standard smoothness and bounded-variance assumptions for stochastic gradients, 
stochastic gradient descent (SGD) with a suitable step-size schedule converges to a first-order stationary point in expectation. 
Formally, if $\nabla_\theta \mathcal{O}$ is $L_{\mathcal{O}}$-Lipschitz and stochastic gradients $\hat g_t$ satisfy 
$\mathbb{E}[\|\hat g_t-\nabla \mathcal{O}(\theta_t)\|^2]\le\sigma^2$, then with constant step size $\eta<1/L_{\mathcal{O}}$ we obtain the standard rate 
$\mathbb{E}\|\nabla \mathcal{O}(\theta_T)\|^2 \le O(1/T)+O(\eta\sigma^2)$. 
In practice, adaptive methods such as Adam used in our experiments empirically achieve similar convergence behavior.
This statement clarifies that the optimization of Equation \ref{eq:composite_equation} yields approximate stationary points of the joint empirical objective.

\subsection{Analysis of surrogate-error propagation}
\label{appendix:surrogate_propagation}

Below we provide an analysis of how surrogate approximation error affects the surrogate-based outcome-consistency loss. This makes explicit the mechanisms by which a nonzero surrogate error can propagate through the inverse model.

Let $\mathbf{l}'\!=\!G_\theta(\mathbf{p},\mathbf{m})$ be a generated pulse for condition $p$, let
\[
r_\phi(\mathbf{l}')\coloneqq S_\phi(\mathbf{l}',\mathbf{p})-\mathbf{m},\qquad r_{\mathrm{true}}(\mathbf{l}')\coloneqq\mathrm{LILAC}(\mathbf{l}',\mathbf{p})-\mathbf{m},
\]
and define the pointwise surrogate error
\[
\varepsilon_\phi(\mathbf{l}')\coloneqq S_\phi(\mathbf{l}',p)-\mathrm{LILAC}(\mathbf{l}',p)=r_\phi(\mathbf{l}')-r_{\mathrm{true}}(\mathbf{l}').
\]

The surrogate-based outcome-consistency term is
\[
\mathcal{L}_S=\|r_\phi(\mathbf{l}')\|^2.
\]
Noting $r_\phi=r_{\mathrm{true}}+\varepsilon_\phi$ we have the exact identity
\[
\mathcal{L}_S=\|r_{\mathrm{true}}(\mathbf{l}')+\varepsilon_\phi(\mathbf{l}')\|^2.
\]
Applying the reverse triangle inequality yields the simple bound
\[
\big|\; \sqrt{\mathcal{L}_S} - \|r_{\mathrm{true}}(\mathbf{l}')\| \;\big|
\le \|\varepsilon_\phi(\mathbf{l}')\|.
\]
Thus, pointwise surrogate error translates directly into an additive bound on the surrogate-implied outcome error: the surrogate loss is a proxy for the true outcome error up to the pointwise surrogate deviation.

\subsection{Surrogate Loss Robustness}

The primary risk of using a surrogate model ($S_\phi$) in the loss function $\mathcal{S}(\theta)$ is that the generator ($G_\theta$) may exploit the surrogate's approximation error $\varepsilon_\phi$ to satisfy the loss function $\mathcal{S}(\theta)$ without satisfying the true physical objective.

\textbf{Conditional Bound}: The generator $G_\theta$ minimizes the true LILAC residual provided that the surrogate pointwise approximation error, $\varepsilon_\phi(\mathbf{l}')\coloneqq S_\phi(\mathbf{l}',\mathbf{p})-\mathrm{LILAC}(\mathbf{l}',\mathbf{p})$, remains uniformly small across the region of laser-pulse space explored by the generator. Using the bound from Section~\ref{appendix:surrogate_propagation}, $\big|\sqrt{\mathcal{L}_S} - \|r_{\mathrm{true}}(\mathbf{l}')\|\big| \le \|\varepsilon_\phi(\mathbf{l}')\|$, the surrogate-implied outcome error and the true LILAC outcome error differ only up to the magnitude of the surrogate approximation error.

\textbf{Adversarial Exploitation}: This robustness can fail if the generator produces an LP, $\mathbf{l}^*$, that acts as an adversarial example for the surrogate. This occurs when $\mathbf{l}^*$ lies in a region of the design space where the surrogate $S_\phi$ is inaccurate, so $\|\varepsilon_\phi(\mathbf{l}^*)\|$ is large. In that case, the generator may achieve low surrogate error $\mathcal{S}(\theta)$ while the true LILAC error remains large, that is, $\|r_{\mathrm{true}}(\mathbf{l}^*)\| \gg 0$. This constitutes exploitation of the surrogate and can produce a physically misleading design. The reconstruction loss $\mathcal{L}(\theta)$ and physics penalty $\mathcal{P}(\theta)$ help mitigate this failure mode by filtering implausible designs. This interpretation is consistent with the ablation results in Section~\ref{sec:fid-div}, where removing $\mathcal{L}(\theta)$ prevents the model from learning physically viable pulses.

 \subsection{Identifiability and the Reconstruction-Weight Trade-off}
 
The ICF LP inverse design problem is a classic example of an ill-posed inverse problem, meaning the solution is non-identifiable: multiple distinct LPs ($\mathbf{l}'_1 \neq \mathbf{l}'_2$) can produce close fusion outcomes ($\mathbf{m}$). 

\textbf{Inverse Problem Constraint}: Minimizing the outcome loss ($\lambda_{\mathcal{S}}\mathcal{S}(\theta)$) alone would force the generator to converge to any LP that satisfies the desired outcome, likely producing an average or non-physical pulse shape located far from the actual manifold of feasible LPs (as described in Section \ref{sec:fid-div}).

\textbf{Regularization by $\lambda_{\mathcal{L}}$}: The Reconstruction Loss ($\lambda_{\mathcal{L}}\mathcal{L}(\theta)$) is required to regularize the non-identifiable problem. By enforcing fidelity to the training data manifold, $\mathcal{L}(\theta)$ forces the generated solution to be the closest physically plausible pulse to the training set that also satisfies the outcome constraint. $\lambda_{\mathcal{L}}$ effectively selects a unique, stable solution from the infinite set of LPs that minimize $\mathcal{S}(\theta)$.

 \subsection{Bias–Variance Trade-offs}

The ablation study varying $(\lambda_{\mathcal{L}}, \lambda_{\mathcal{S}})$ (Figure \ref{tab:metrics}) explicitly maps the bias–variance trade-off in the generated output:

\textbf{Variance ($\lambda_{\mathcal{S}}$ dominance)}: A low $\lambda_{\mathcal{L}}$ encourages the generator to explore further from the training manifold to satisfy the outcome, thus leading to high diversity and an increased variance.

\textbf{Bias ($\lambda_{\mathcal{L}}$ dominance)}: A high $\lambda_{\mathcal{L}}$ severely constrains the generator to the known training data, leading to low diversity (low variance). This strategy is robust but prone to bias against novel pulse shapes that may lie slightly outside the training data distribution.

\section{Correlation between outcome-consistency error and out-of-distribution data}

\label{appendix:surrogate_novelty}

A key concern when employing surrogate models is whether their approximation error remains stable when the generator produces pulses that differ from those seen during surrogate training. To quantify this effect, we compare the implosion-outcome error for generated pulses with a simple novelty measure computed relative to the original training dataset.

\paragraph{Novelty score.}
Given a generated pulse $\mathbf{l}'$, we define its novelty as its distance
to the closest pulse in the training set $\mathcal{D}_{\mathrm{train}}$:
\[
d(\mathbf{l}')
=
\min_{\mathbf{l}\in \mathcal{D}_{\mathrm{train}}} \| \mathbf{l} - \mathbf{l}' \|_2.
\]
This quantity measures how far the generator has moved away from the region where
the surrogate has been trained and validated, and therefore acts as a proxy for
out-of-distribution (OOD) deviation.

\paragraph{Implosion-Outcome Error.}
We compute the surrogate outcome-consistency error
\[
e(\mathbf{l}') = \left\| \mathbf{m} - \mathbf{S}_\phi(\mathbf{l}', \mathbf{p}) \right\|^2.
\]
The magnitude $e(\mathbf{l}')$ reflects the squared surrogate residual induced by $\mathbf{S}_\phi$. This is the same quantity used inside the loss in Equation \ref{eq:surrogate_loss_in_generator}; unlike the main-text evaluation metric, it is not reported as MAPE.

\paragraph{Correlation analysis.}
To assess whether $e(\mathbf{l}')$ systematically increases as pulses move farther
from the training manifold, we compute the Spearman rank correlation
\[
\rho_{\mathrm{Spearman}}
=
\operatorname{corr}_{\mathrm{Spearman}}( \{d(\mathbf{l}')\}, \{e(\mathbf{l}')\} ).
\]
Spearman correlation is insensitive to scale and captures monotonic relationships, making it appropriate for detecting trends such as ``greater novelty $\Rightarrow$ larger surrogate error.'' A strong positive correlation would indicate that $\mathbf{S}_\phi$ introduces a strong bias in the generative model against OOD regions, while a correlation close to zero would suggest that the surrogate maintains relatively uniform influence across the generator's operating domain.

We ran this evaluation using Diffusion as the base model since it offers a good balance between $\mathbf{m}$ error and diversity. The Spearman correlation between implosion-outcome error magnitude and the novelty score on the test set is $\mathbf{\rho_{\mathrm{Spearman}}=0.353}$ with $p<10^{-290}$. This modest correlation indicates that the surrogate error
exhibits a mild monotonic increase with distance from the training manifold.
The effect size is not strong, suggesting that the surrogate remains reasonably reliable across the region of pulse space that the generative
models operate in. Importantly, this behavior is consistent with expectations for neural surrogates trained on physically coherent waveform datasets: their
accuracy degrades smoothly with novelty but does not break down or dramatically inflate in regions relevant to our experiments.

Moreover, as shown by the diffusion model results, diversity does not negatively affect $\mathbf{m}$, since the model achieves high diversity while maintaining low $\mathbf{m}$ error.

\paragraph{Distance of generated pulses from the training manifold.}
The correlation above shows that surrogate error grows only mildly with novelty; we further check how far generated pulses actually move from the training data. We compute the novelty score $d(\cdot)$ for three populations: training pulses (distance to their nearest other training pulse, $N{=}100{,}000$), test pulses ($N{=}692{,}700$), and pulses generated by $\mathbf{LPDS}_\text{Diff}$ for one of the held-out pellet clusters of Appendix~\ref{sec:pellet_stratified} ($N{=}5{,}000$). Table~\ref{tab:novelty_populations} reports the mean novelty of each population and the fraction of samples beyond the 95th percentile of the test-set novelty distribution.

\begin{table}[h!]
\centering
\begin{tabular}{lccc}
\toprule
\textbf{Population} & \textbf{Mean novelty} & \textbf{95th percentile} & \textbf{Fraction beyond test 95th percentile} \\
\midrule
Training (nearest other training pulse) & 0.7125 & 1.0807 & -- \\
Test & 0.7125 & 1.0814 & 5.00\% \\
$\mathbf{LPDS}_\text{Diff}$ (held-out pellet cluster) & 0.7048 & -- & 5.04\% \\
\bottomrule
\end{tabular}
\caption{Novelty of generated pulses relative to the training set, compared with held-out test pulses.}
\label{tab:novelty_populations}
\end{table}

Generated pulses are, on average, slightly closer to the training data than held-out test pulses ($0.7048$ vs.\ $0.7125$), and the fraction beyond the test 95th percentile ($5.04\%$) matches the $5\%$ expected for in-distribution samples. Together with the mild correlation reported above, this indicates that $\mathbf{LPDS}_\text{Diff}$ neither drifts away from the training manifold nor relies on off-manifold regions where the surrogate is unreliable.

\section{Inpainting}

To enhance controllability during generation, LPDS supports inpainting as a mechanism for partial conditioning directly in the laser pulse space. This allows domain experts to specify only certain segments of a laser pulse, while leaving the remaining parts to be completed by the model in a physically consistent manner.

Formally, let $\mathbf{l} \in \mathbb{R}^T$ denote a laser pulse of length $T$, and let $\mathbf{b} \in \{0,1\}^T$ be a binary mask that defines the constraint region. Each element $\mathbf{b}_t$ indicates whether the $t$-th component of the pulse is observed ($\mathbf{b}_t = 1$) or masked ($\mathbf{b}_t = 0$). Let $\mathbf{l}_{\text{obs}}$ represent the observed (non-masked) entries of the pulse.

To incorporate inpainting with Diffusion, we modify the reverse diffusion process so that known regions remain fixed at each step. Let $\hat{\mathbf{l}}_n$ be the predicted intermediate pulse at timestep $n$ from the standard DDIM update rule (Equation~\ref{equation:l_forward}). We define the inpainted latent state as:
\begin{equation}
\mathbf{l}_n^{\text{inpaint}} = \mathbf{b} \odot \mathbf{l}_{\text{obs}} + (1 - \mathbf{b}) \odot \hat{\mathbf{l}}_n,
\end{equation}
where $\odot$ denotes elementwise multiplication. This formulation ensures that the observed parts of the pulse are preserved throughout the denoising process, while the model only modifies the masked (unknown) components.

In practice, the mask $\mathbf{b}$ can be flexibly defined to support a wide range of expert interactions, for example, preserving a fixed peak intensity, pre-specified rising edge, or constraints learned from past experiments. Combined with the implosion outcomes and target parameters conditioning via $\mathbf{m}$ and $\mathbf{p}$, inpainting enables LPDS to support fine-grained, interpretable, and editable pulse design.


\begin{figure}[t]
    \centering
    \begin{subfigure}[b]{0.48\textwidth}
        \centering
        \includegraphics[width=\columnwidth]{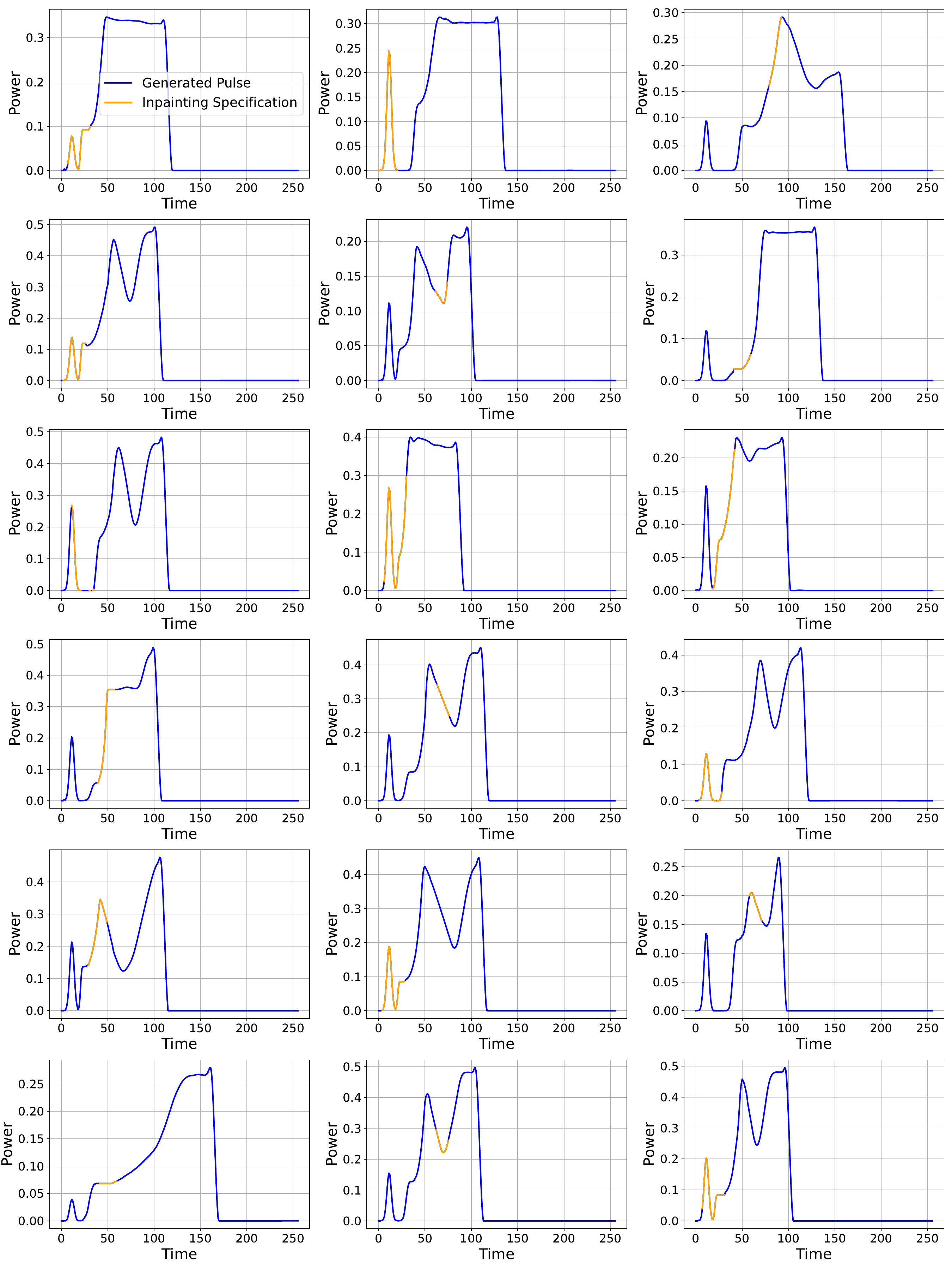}
        \caption{LPs generated by the model when doing inpainting using Diffusion.}
        \label{figure:inpainting_lps}
    \end{subfigure}
    \hfill
    \begin{subfigure}[b]{0.48\textwidth}
        \centering
        \includegraphics[width=\columnwidth]{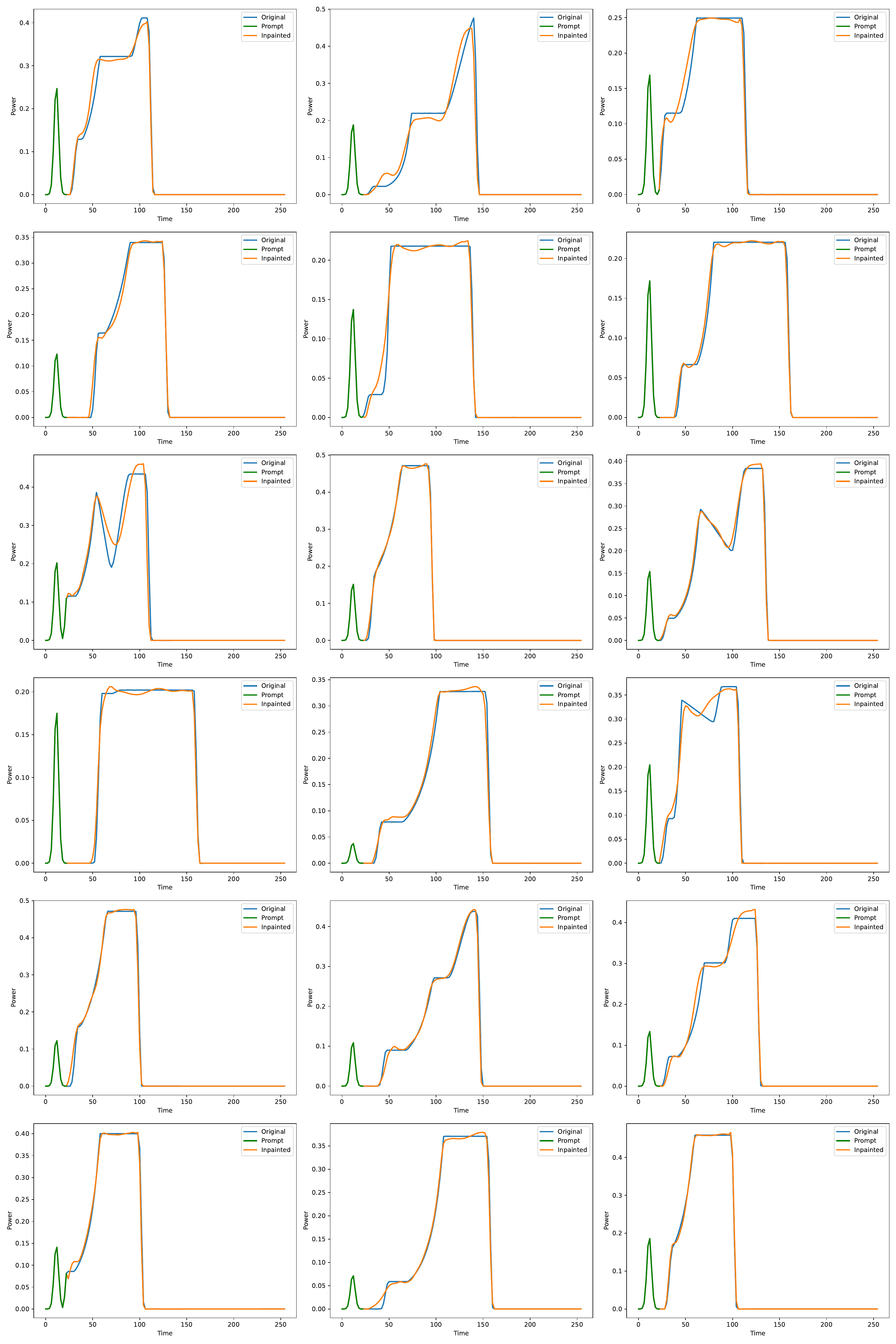}
        \caption{LPs generated by the model when doing inpainting using the LSTM.}
        \label{figure:inpainting_ar_lps}
    \end{subfigure}
\end{figure}

\section{Example Pulse Shape Generation}
\label{section:example pulse shape generation}

\begin{figure}[t]
    \centering
    \begin{subfigure}[b]{0.48\textwidth}
        \centering
        \includegraphics[width=\textwidth]{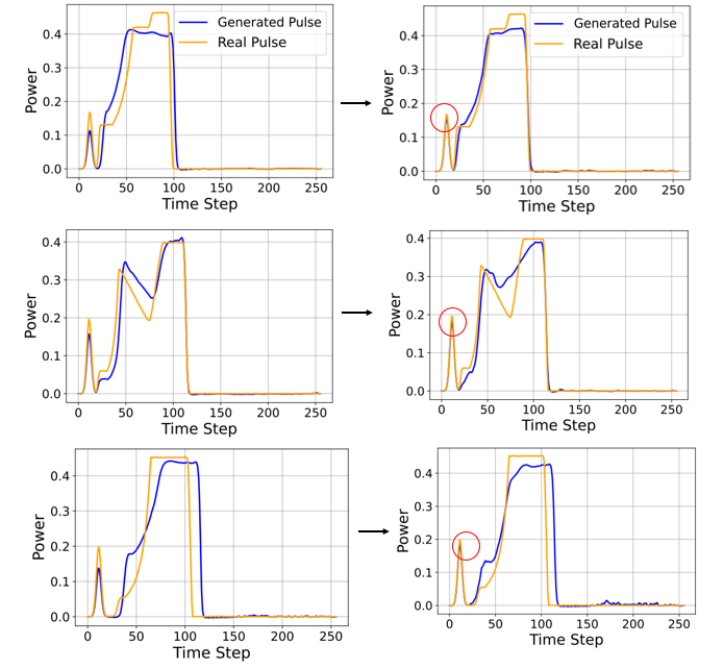}
        \caption{LPs constraining on Picket Power (red highlight). The model successfully constrains the picket power to the target value while allowing variation in the rest of the LP.}
        \label{figure:pulse_constrain_diffusion}
    \end{subfigure}
    \hfill
    \begin{subfigure}[b]{0.48\textwidth}
        \centering
        \includegraphics[width=\textwidth]{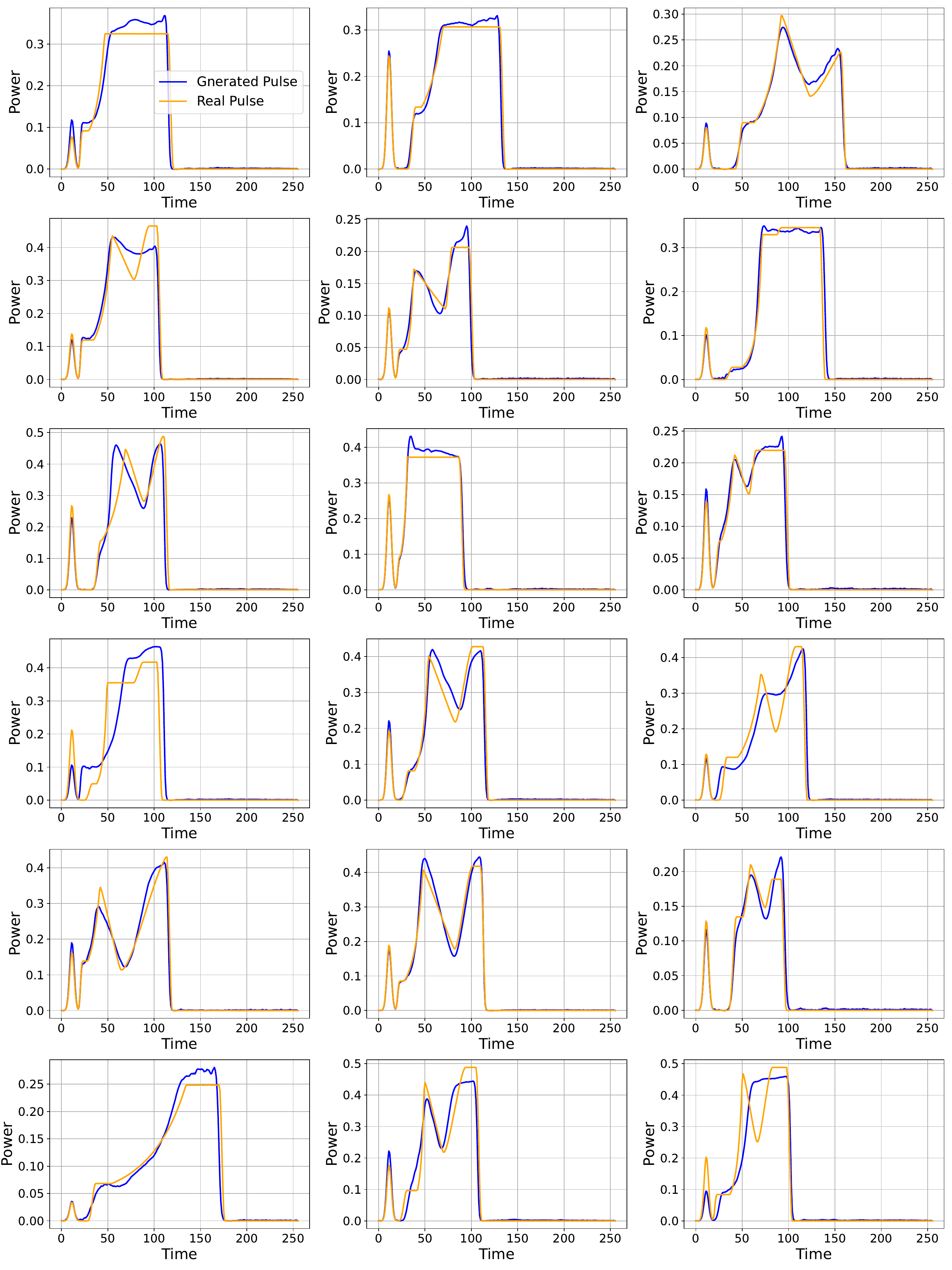}
        \caption{LPs Generated by Diffusion.}
        \label{figure:ps_ours}
    \end{subfigure}

\end{figure}

\begin{figure}[t]
    \begin{subfigure}[b]{0.5\textwidth}
        \centering
        \includegraphics[width=\textwidth]{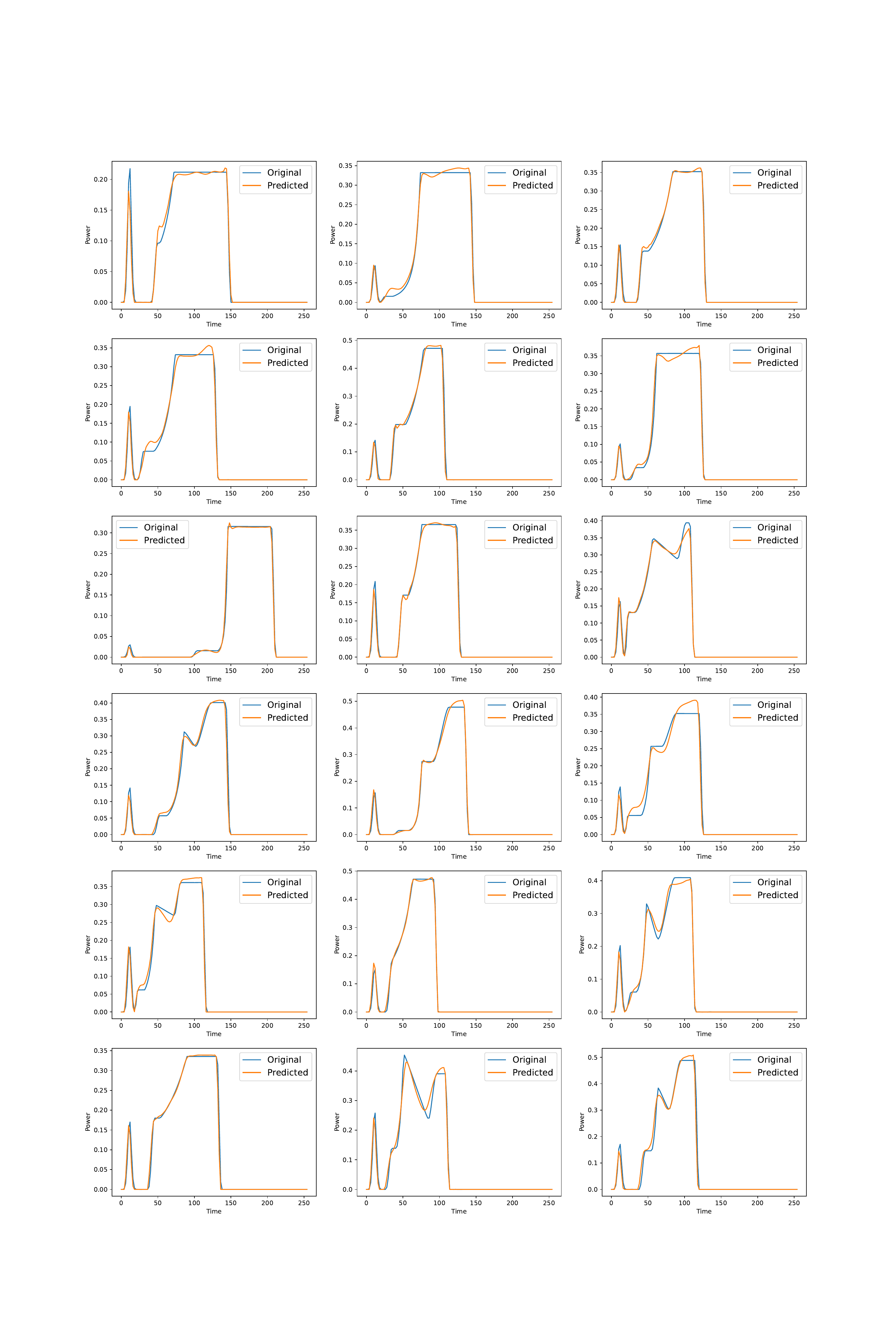}
        \caption{LPs generated by  LSTM  model}
        \label{figure:ps-lstm}
    \end{subfigure}
    \begin{subfigure}[b]{0.5\textwidth}
        \centering
        \includegraphics[width=\textwidth]{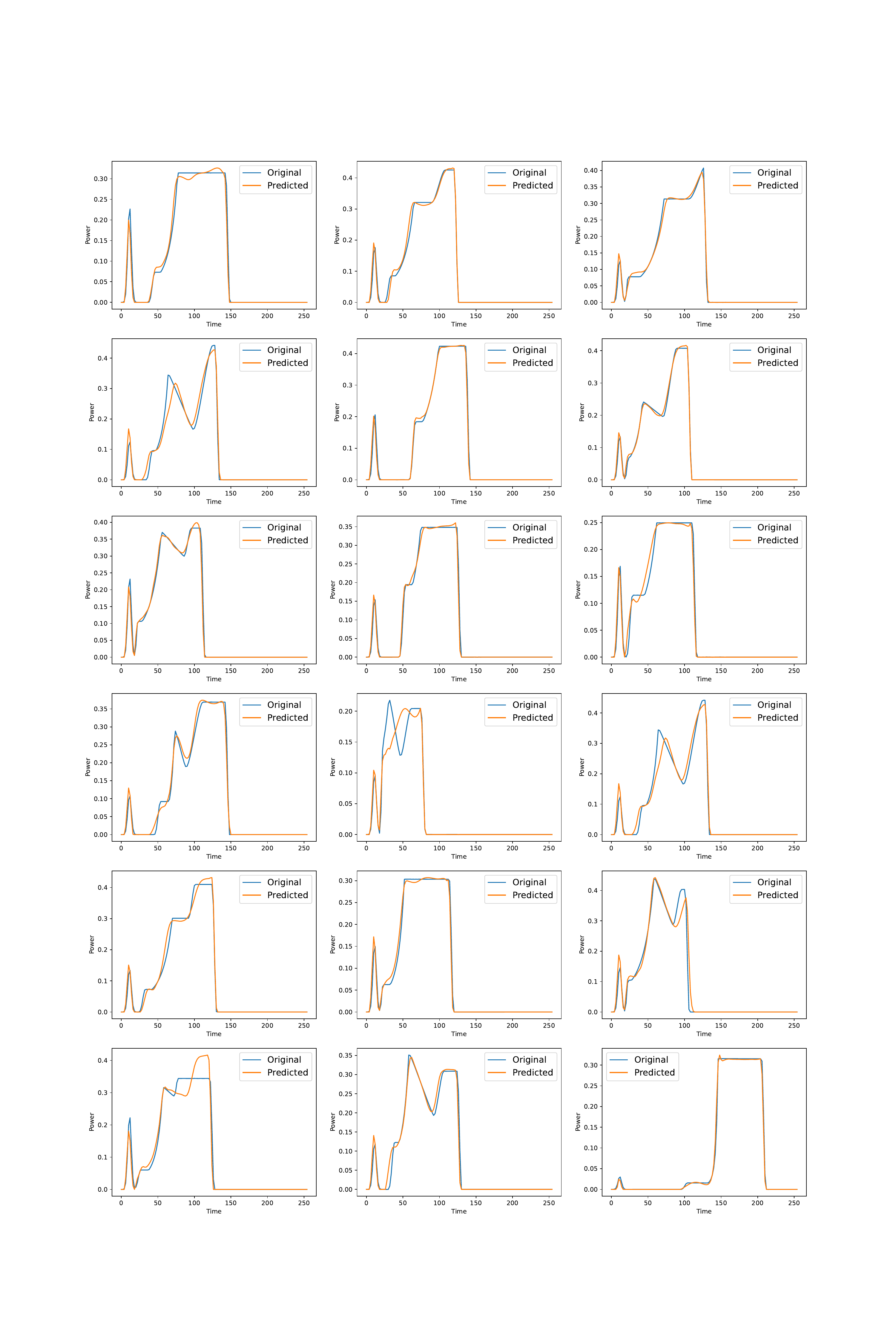}
        \caption{LPs generated by the  Transformer  model}
        \label{figure:ps-transformer}
    \end{subfigure}
    \caption{LPs generated by predictive autoregressive models.}
\end{figure}

\begin{figure}[t]
    \begin{subfigure}[b]{0.5\textwidth}
        \centering
        \includegraphics[width=\textwidth]{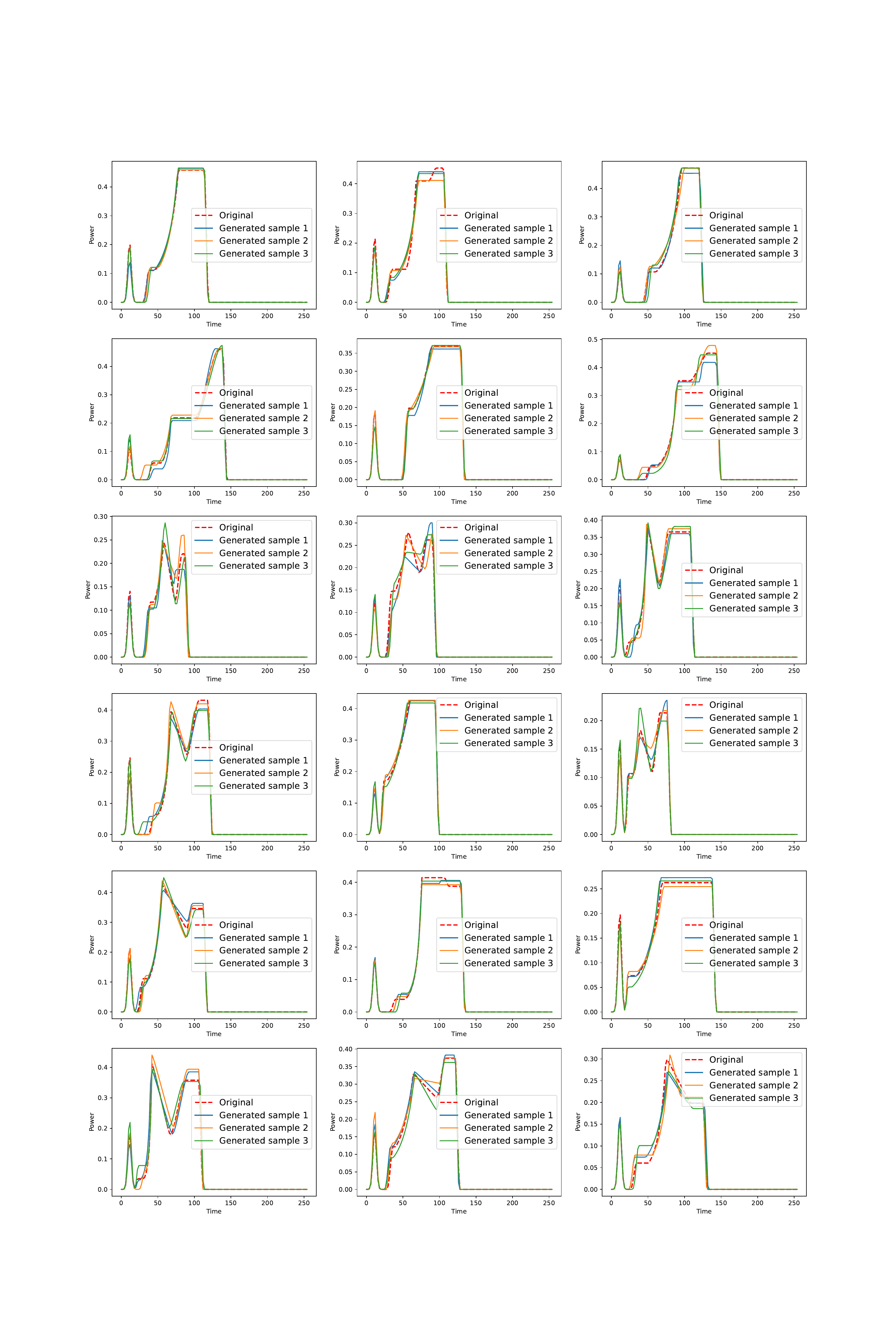}
        \caption{LPs generated by  GaussianAR  model}
        \label{figure:ps-guassian}
    \end{subfigure}
    \begin{subfigure}[b]{0.5\textwidth}
        \centering
        \includegraphics[width=\textwidth]{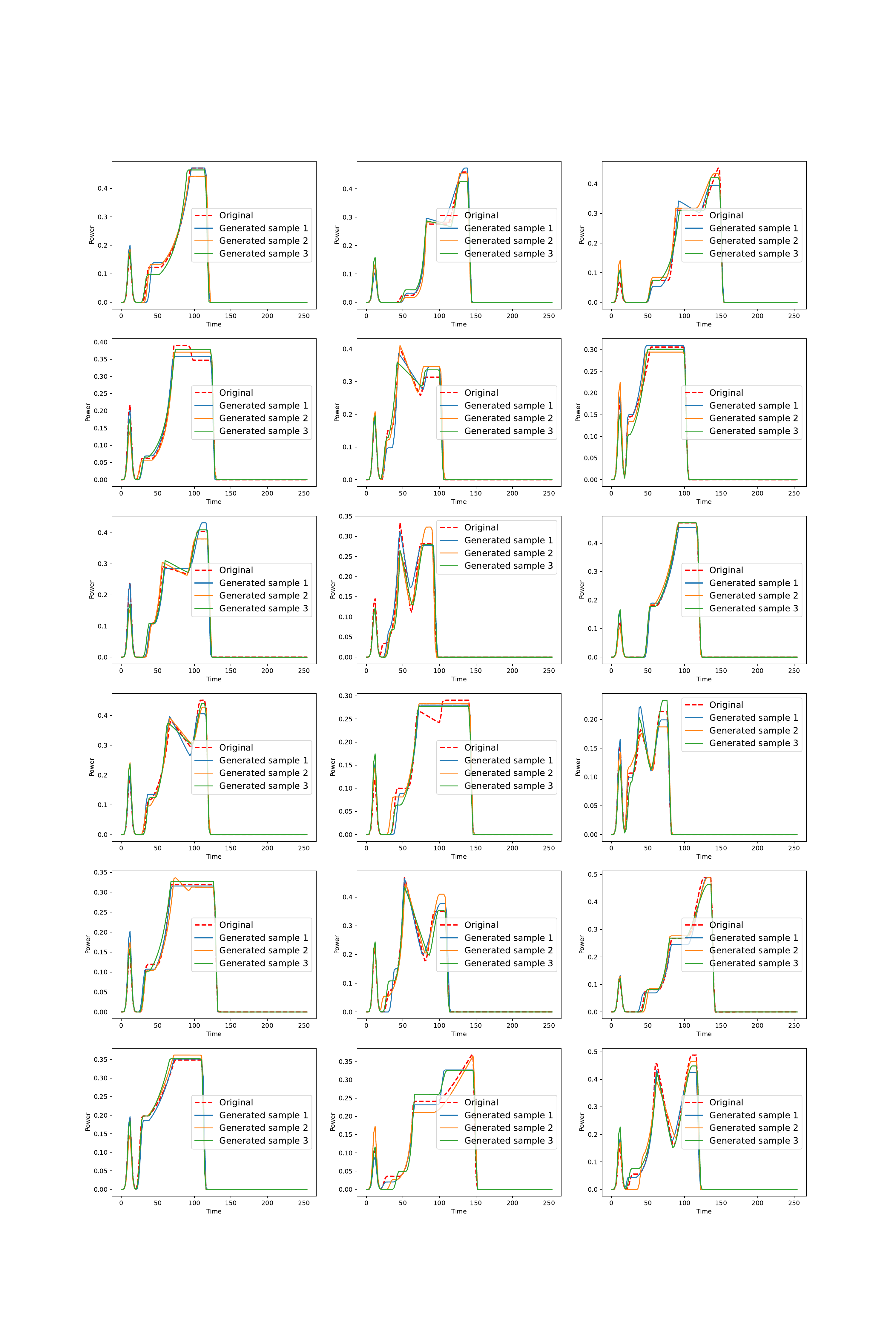}
        \caption{LPs generated by the  MixtureofGaussianAR  model}
        \label{figure:ps-mixture_gaussians}
    \end{subfigure}
    \caption{LPs generated by generative autoregressive model}
\end{figure}

\begin{figure}[t]
    \centering
        \begin{subfigure}[b]{0.48\textwidth}
        \centering
        \includegraphics[width=\textwidth]{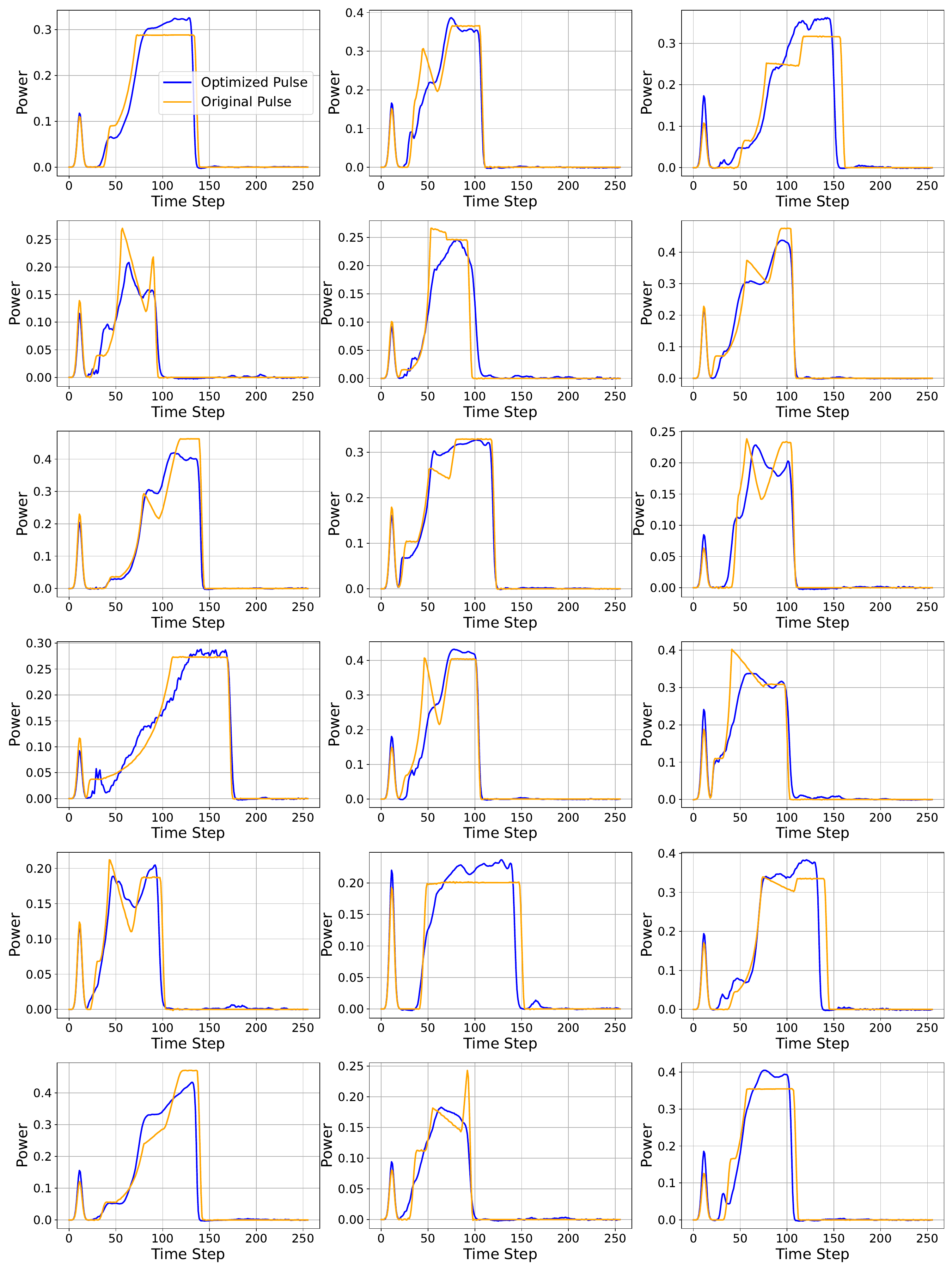}
        \caption{LPs Generated by $\mathbf{LPDS}_\text{Diff}$ after optimizing for enhancing energy yield.}
        \label{figure:ps_optimization}
    \end{subfigure}
    \hfill
    \begin{subfigure}[b]{0.48\textwidth}
        \centering
        \includegraphics[width=\textwidth]{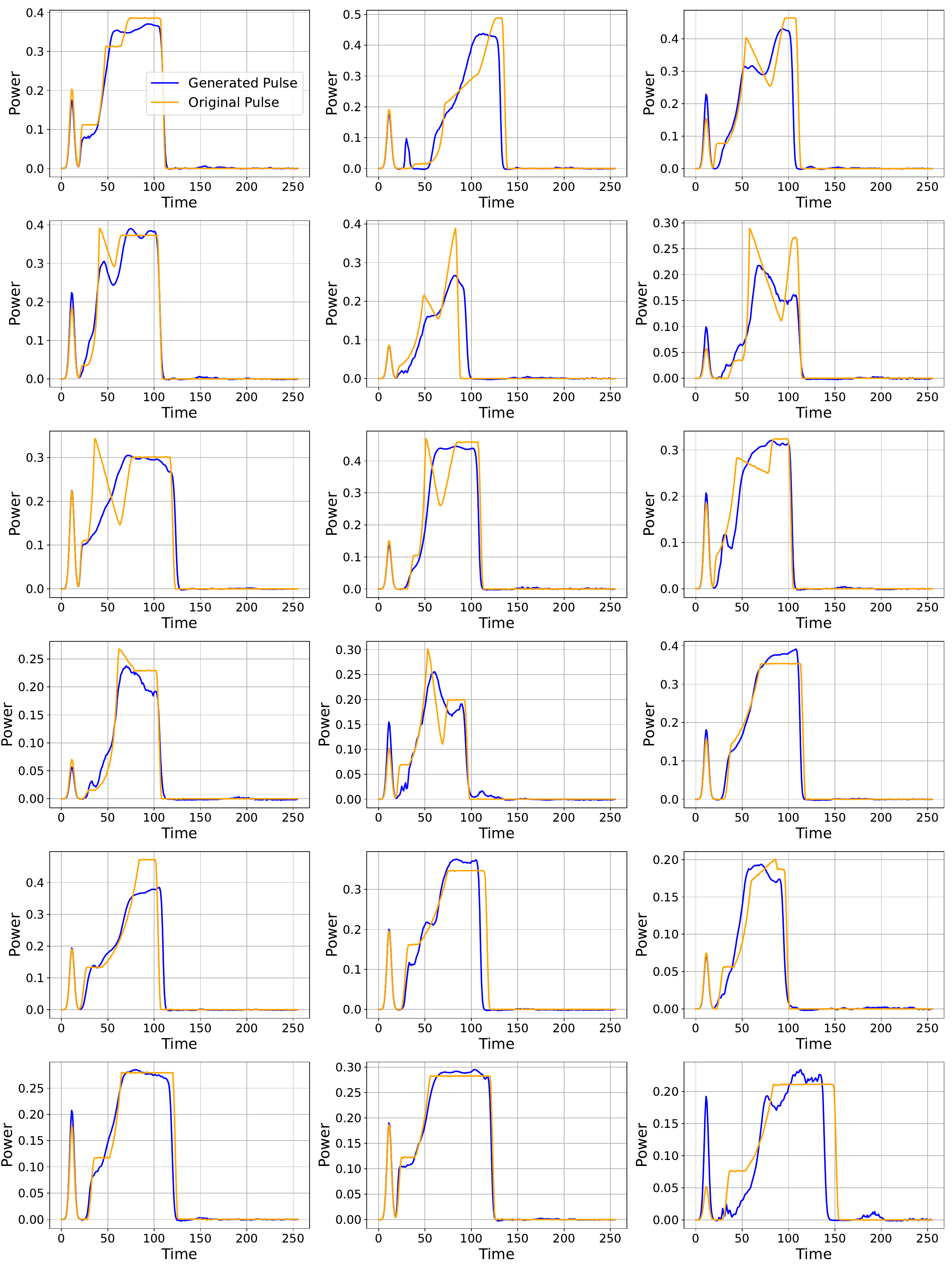}
        \caption{LPs Generated by $\mathbf{LPDS}_\text{Diff}$ after optimizing for enhancing areal density.}
        \label{figure:ps_optimization_rhoar}
    \end{subfigure}
    \caption{Examples of LPs generated by $\mathbf{LPDS}_\text{Diff}$ with the outcome-optimization procedure described in Section \ref{section:optimization}.}
    \label{figure:ps_optimization_lp}
\end{figure}

We present some of the pulses generated by the different approaches. Figure \ref{figure:ps_ours} for Diffusion, Figure \ref{figure:ps-lstm} for LSTM, Figure \ref{figure:ps-transformer} for Transformer, Figure \ref{figure:ps-guassian} for GaussianAR, Figure \ref{figure:ps-mixture_gaussians} for MixtureOfGaussianAR. Figure \ref{figure:ps_optimization_lp} shows examples of LPs generated by the LP novel design method described in Section \ref{section:optimization}


\section{Broader Impacts}
This work aims to accelerate inverse design for inertial confinement fusion, which could contribute positively to scientific discovery in high-energy-density physics and, in the long term, to cleaner energy technologies if fusion becomes practical. More broadly, the methodological contributions may also benefit other scientific inverse-design problems where multiple physically plausible solutions must be generated under hard constraints.

Potential negative impacts are primarily indirect. Improved inverse-design and optimization tools could be misapplied to accelerate the design of high-energy systems beyond the intended scientific context.

\clearpage

\input{checklist.tex}

\end{document}

%% file: checklist.tex
\section*{NeurIPS Paper Checklist}

\begin{enumerate}

\item {\bf Claims}
    \item[] Question: Do the main claims made in the abstract and introduction accurately reflect the paper's contributions and scope?
    \item[] Answer: \answerYes{}
    \item[] Justification: Yes. The abstract and introduction state the main claims about LPDS, its controllable generation mechanisms, and simulator-grounded validation, and these claims are supported by the experimental and theoretical sections.

\item {\bf Limitations}
    \item[] Question: Does the paper discuss the limitations of the work performed by the authors?
    \item[] Answer: \answerYes{}
    \item[] Justification: Yes. The paper includes a dedicated Limitations section discussing regime.

\item {\bf Theory assumptions and proofs}
    \item[] Question: For each theoretical result, does the paper provide the full set of assumptions and a complete (and correct) proof?
    \item[] Answer: \answerNA{}
    \item[] Justification: The paper does not include theoretical results

    \item {\bf Experimental result reproducibility}
    \item[] Question: Does the paper fully disclose all the information needed to reproduce the main experimental results of the paper to the extent that it affects the main claims and/or conclusions of the paper (regardless of whether the code and data are provided or not)?
    \item[] Answer: \answerYes{}
    \item[] Justification: Yes. The paper specifies the dataset construction, evaluation metrics, number of runs and seeds, train/test splits, hyperparameter search ranges, optimizer choices, and compute setup in the main paper and appendix Experimental Details.

\item {\bf Open access to data and code}
    \item[] Question: Does the paper provide open access to the data and code, with sufficient instructions to faithfully reproduce the main experimental results, as described in supplemental material?
    \item[] Answer: \answerNo{}
    \item[] Justification: We will release the code upon publication. 

\item {\bf Experimental setting/details}
    \item[] Question: Does the paper specify all the training and test details (e.g., data splits, hyperparameters, how they were chosen, type of optimizer) necessary to understand the results?
    \item[] Answer: \answerYes{}
    \item[] Justification: Yes. The paper and appendix specify the data split, training setup, hyperparameter ranges, selected hyperparameters, optimizer, scheduler, and model-specific settings needed to understand the experiments.

\item {\bf Experiment statistical significance}
    \item[] Question: Does the paper report error bars suitably and correctly defined or other appropriate information about the statistical significance of the experiments?
    \item[] Answer: \answerYes{}
    \item[] Justification: Yes. The main result tables report variability with $\pm$ values.

\item {\bf Experiments compute resources}
    \item[] Question: For each experiment, does the paper provide sufficient information on the computer resources (type of compute workers, memory, time of execution) needed to reproduce the experiments?
    \item[] Answer: \answerYes{}
    \item[] Justification: Yes. The appendix reports the hardware used, total project-scale compute, single-GPU timing assumptions, average training times, throughput, and latency measurements for the main approaches.
    
\item {\bf Code of ethics}
    \item[] Question: Does the research conducted in the paper conform, in every respect, with the NeurIPS Code of Ethics \url{https://neurips.cc/public/EthicsGuidelines}?
    \item[] Answer: \answerYes{}
    \item[] Justification: Yes. To the best of our knowledge, the research described in the paper conforms to the NeurIPS Code of Ethics.

\item {\bf Broader impacts}
    \item[] Question: Does the paper discuss both potential positive societal impacts and negative societal impacts of the work performed?
    \item[] Answer: \answerYes{}
    \item[] Justification: Yes. The appendix now includes a dedicated Broader Impacts section discussing potential positive scientific and energy-related benefits as well as possible misuse risks.
    
\item {\bf Safeguards}
    \item[] Question: Does the paper describe safeguards that have been put in place for responsible release of data or models that have a high risk for misuse (e.g., pre-trained language models, image generators, or scraped datasets)?
    \item[] Answer: \answerNA{}
    \item[] Justification: The paper does not release high-risk generative models or scraped datasets.

\item {\bf Licenses for existing assets}
    \item[] Question: Are the creators or original owners of assets (e.g., code, data, models), used in the paper, properly credited and are the license and terms of use explicitly mentioned and properly respected?
    \item[] Answer: \answerNA{}
    \item[] Justification: We do not use existing assets.

\item {\bf New assets}
    \item[] Question: Are new assets introduced in the paper well documented and is the documentation provided alongside the assets?
    \item[] Answer: \answerNA{}
    \item[] Justification: The paper does not release new assets.

\item {\bf Crowdsourcing and research with human subjects}
    \item[] Question: For crowdsourcing experiments and research with human subjects, does the paper include the full text of instructions given to participants and screenshots, if applicable, as well as details about compensation (if any)? 
    \item[] Answer: \answerNA{}
    \item[] Justification: The work does not involve crowdsourcing or research with human subjects.

\item {\bf Institutional review board (IRB) approvals or equivalent for research with human subjects}
    \item[] Question: Does the paper describe potential risks incurred by study participants, whether such risks were disclosed to the subjects, and whether Institutional Review Board (IRB) approvals (or an equivalent approval/review based on the requirements of your country or institution) were obtained?
    \item[] Answer: \answerNA{}
    \item[] Justification: The work does not involve crowdsourcing or research with human subjects.

\item {\bf Declaration of LLM usage}
    \item[] Question: Does the paper describe the usage of LLMs if it is an important, original, or non-standard component of the core methods in this research? Note that if the LLM is used only for writing, editing, or formatting purposes and does \emph{not} impact the core methodology, scientific rigor, or originality of the research, declaration is not required.
    \item[] Answer: \answerNA{}
    \item[] Justification: The core method does not use LLMs as an important, original, or non-standard methodological component.

\end{enumerate}